%% file: main.tex
\documentclass[11pt]{article}

\usepackage[english]{babel}
\usepackage[letterpaper,margin=1in]{geometry}

\input{preamble}

\begin{document}
\title{A Modality-agnostic Multi-task Foundation Model for\\Human Brain Imaging}

\author{Peirong Liu, Oula Puonti, Xiaoling Hu, Karthik Gopinath, Annabel Sorby-Adams\\Daniel C. Alexander, W. Taylor Kimberly and Juan E. Iglesias
}

\maketitle

\blfootnote{This
work was primarily supported by the NIH Grant 1RF1AG080371 (Principal investigator: Juan E. Iglesias). Additional support from NIH 1RF1AG080371, 1R01EB031114, 1R01AG070988, 1RF1MH123195. ASA is funded by the American Heart Association Postdoctoral Fellowship and the Fulbright Commission. OP is supported by Lundbeckfonden R360–2021–395. Wellcome Trust award 221915/Z/20/Z and the NIHR UCLH Biomedical Research Centre support DCA on this topic.}
\blfootnote{Peirong Liu (e-mail: peirong@jhu.edu) is with the Department of Electrical and Computer Engineering at Johns Hopkins University, Baltimore, MD 21218, USA.}
\blfootnote{Xiaoling Hu, Karthik Gopinath, Annabel Sorby-Adams, W. Taylor Kimberly, and Juan E. Iglesias are with the Athinoula A. Martinos Center for Biomedical Imaging, Harvard Medical School and Massachusetts General Hospital, Boston, MA 02129, USA.}
\blfootnote{Oula Puonti is with the Danish Research Centre for Magnetic Resonance, 2650 Hvidovre, Denmark.}
\blfootnote{Daniel C. Alexander is with the Centre for Medical Image Computing, University College London, London WC1E 6BT, United Kingdom.}

\input{sec/abs}
\input{sec/intro}

\input{sec/related_work}

\input{sec/method/main}

\input{sec/exp/main}
\input{sec/con}

\bibliographystyle{IEEEtran}
\bibliography{reference}

\end{document}

%% file: preamble.tex
\providecommand{\keywords}[1]
{
  \small	
  \textbf{\textit{Keywords---}} #1
}

\usepackage{cite}

\usepackage{amsmath}
\usepackage{amssymb,amsfonts}
\usepackage{algorithmic}
\usepackage{graphicx}
\usepackage{textcomp}

\usepackage{amsthm} 
\usepackage{pifont}
\newcommand{\cmark}{\ding{51}}%
\newcommand{\xmark}{\ding{55}}%

\usepackage{booktabs}
\usepackage{graphicx}
\usepackage{wrapfig}

\usepackage[pagebackref,breaklinks,colorlinks]{hyperref}
\usepackage[capitalize]{cleveref}
\crefname{section}{Sec.}{Secs.}
\Crefname{section}{Section}{Sections}
\Crefname{table}{Table}{Tables}
\crefname{table}{Tab.}{Tabs.}

\usepackage{xcolor}
\usepackage{tikz}
\usepackage{changepage}
\usepackage{tikz-3dplot}
\usepackage{tikzscale}
\usetikzlibrary{plotmarks}
\usepackage{pgfplots}
\usetikzlibrary{fadings}
\usetikzlibrary{tikzmark,calc} 
\usetikzlibrary{shapes.arrows}
\usetikzlibrary{decorations.pathreplacing, decorations.markings}
\usepgfplotslibrary{colorbrewer}
\usepgfmodule{nonlineartransformations} 
\makeatletter
\def\latticetilt{%
\pgf@xa=\pgf@x%
\pgf@ya=\pgf@y%
\pgfmathsetmacro{\myx}{\pgf@xa+\pgfkeysvalueof{/tikz/lattice/amplitude}*sin((\pgf@ya/\pgfkeysvalueof{/tikz/lattice/spacing})*360/\pgfkeysvalueof{/tikz/lattice/superlattice period})}%
\pgf@x=\myx pt%
\pgfmathsetmacro{\myy}{\pgf@ya+\pgfkeysvalueof{/tikz/lattice/amplitude}*sin((\pgf@xa/\pgfkeysvalueof{/tikz/lattice/spacing})*360/\pgfkeysvalueof{/tikz/lattice/superlattice period})}%
\pgf@y=\myy pt}

\usepackage[capbesideposition=right]{floatrow}

\xdefinecolor{navyblue}{RGB}{19 41 75}
\xdefinecolor{carolinablue}{RGB}{123 175 212}
\xdefinecolor{cb}{RGB}{123 175 212}
\xdefinecolor{flatgrey}{RGB}{162 170 173}
\xdefinecolor{matcha}{RGB}{183 224 176}

\definecolor{mygreen}{HTML}{DCEDC8}
\definecolor{lightgrey}{HTML}{E1E1E1}
\definecolor{darkblue}{HTML}{00275D}
\definecolor{mint}{HTML}{99EDC3}

\xdefinecolor{myyellow}{HTML}{FFE082}
\xdefinecolor{myblue}{RGB}{181, 222, 255}
\xdefinecolor{mypurple}{RGB}{202, 184, 255}
\xdefinecolor{myorange}{RGB}{255, 171, 145}
\xdefinecolor{tomato}{RGB}{255, 99, 71}

\definecolor{applegreen}{rgb}{0.4, 0.81, 0.0}
\definecolor{bittersweet}{rgb}{1.0, 0.2, 0.3}
\definecolor{mayablue}{rgb}{0.4, 0.5, 0.98}
\definecolor{citrine}{rgb}{1, 0.62, 0.14}
\definecolor{bubblegum}{rgb}{0.99, 0.76, 0.8}
\definecolor{olive}{rgb}{0.42, 0.56, 0.14}

\xdefinecolor{mgh}{RGB}{0 139 176}
\xdefinecolor{harvardcrimson}{RGB}{165 28 48}
\xdefinecolor{hc}{RGB}{165 28 48} 
\xdefinecolor{harvardred}{RGB}{237 27 52}
\xdefinecolor{hr}{RGB}{237 27 52}
\xdefinecolor{harvardsalmon}{RGB}{236 143 156}
\xdefinecolor{hs}{RGB}{165 28 48}
\xdefinecolor{harvardblack}{RGB}{30 30 30}
\xdefinecolor{hblack}{RGB}{30 30 30}
\xdefinecolor{harvardgrey}{RGB}{147 161 173}
\xdefinecolor{hgrey}{RGB}{147 161 173}
\xdefinecolor{harvardgreen}{RGB}{77 184 72}
\xdefinecolor{hgreen}{RGB}{77 184 72}
\xdefinecolor{harvardlime}{RGB}{203 219 42}
\xdefinecolor{hlime}{RGB}{203 219 42}
\xdefinecolor{harvardblue}{RGB}{78 136 199}
\xdefinecolor{hblue}{RGB}{78 136 199}
\xdefinecolor{harvardsky}{RGB}{149 181 223}
\xdefinecolor{hsky}{RGB}{149 181 223}
\xdefinecolor{harvardwarmyellow}{RGB}{252 179 21}
\xdefinecolor{hwy}{RGB}{252 179 21}
\xdefinecolor{harvardyellow}{RGB}{255 222 45}
\xdefinecolor{hy}{RGB}{255 222 45}
\xdefinecolor{harvardturq}{RGB}{0 170 173}
\xdefinecolor{ht}{RGB}{0 170 173}
\xdefinecolor{harvardaqua}{RGB}{80 196 221}
\xdefinecolor{ha}{RGB}{80 196 221}
\xdefinecolor{harvardpurple}{RGB}{148 110 183}
\xdefinecolor{hp}{RGB}{148 110 183}
\xdefinecolor{harvardlavender}{RGB}{187 137 202}
\xdefinecolor{hl}{RGB}{187 137 202}

\newcommand{\dashedLayer}[6]{
			\def\a{#1} 
			\def\b{0.02}
			\def\c{#2} 
			\def\t{#3} 
			\def\d{#4} 

			\draw[line width=0.15mm, dash pattern=on 2.25pt off 1.8pt](\c+\t,0,\d) -- (\c+\t,\a,\d) -- (\t,\a,\d);                                                      
			\draw[line width=0.15mm, dash pattern=on 2.25pt off 1.8pt](\t,0,\a+\d) -- (\c+\t,0,\a+\d) node[midway,below] {#6} -- (\c+\t,\a,\a+\d) -- (\t,\a,\a+\d) -- (\t,0,\a+\d); 
			\draw[line width=0.15mm, dash pattern=on 2.25pt off 1.8pt](\c+\t,0,\d) -- (\c+\t,0,\a+\d);
			\draw[line width=0.15mm, dash pattern=on 2.25pt off 1.8pt](\c+\t,\a,\d) -- (\c+\t,\a,\a+\d);
			\draw[line width=0.15mm, dash pattern=on 2.25pt off 1.8pt](\t,\a,\d) -- (\t,\a,\a+\d);
			
			\draw[line width=0.15mm] (\c+\t,0,\d) -- (\c+\t,\a,\d);
			\draw[line width=0.15mm] (\c+\t,0,\d) -- (\c+\t,0,\a+\d);
			\draw[line width=0.15mm] (\c+\t,\a,\d) -- (\c+\t,\a,\a+\d);
			\draw[line width=0.15mm] (\c+\t,0,\a+\d) -- (\c+\t,\a,\a+\d);
			
			\filldraw[#5] (\t+\b,\b,\a+\d) -- (\c+\t-\b,\b,\a+\d) -- (\c+\t-\b,\a-\b,\a+\d) -- (\t+\b,\a-\b,\a+\d) -- (\t+\b,\b,\a+\d); 
			\filldraw[#5] (\t+\b,\a,\a-\b+\d) -- (\c+\t-\b,\a,\a-\b+\d) -- (\c+\t-\b,\a,\b+\d) -- (\t+\b,\a,\b+\d);
			\ifthenelse {\equal{#5} {}}
			{} 
			{\filldraw[#5] (\c+\t,\b,\a-\b+\d) -- (\c+\t,\b,\b+\d) -- (\c+\t,\a-\b,\b+\d) -- (\c+\t,\a-\b,\a-\b+\d);} 
		}

\newcommand{\networkLayer}[6]{
			\def\a{#1} 
			\def\b{0.02}
			\def\c{#2} 
			\def\t{#3} 
			\def\d{#4} 

			\draw[line width=0.15mm](\c+\t,0,\d) -- (\c+\t,\a,\d) -- (\t,\a,\d);                                                      
			\draw[line width=0.15mm](\t,0,\a+\d) -- (\c+\t,0,\a+\d) node[midway,below] {#6} -- (\c+\t,\a,\a+\d) -- (\t,\a,\a+\d) -- (\t,0,\a+\d); 
			\draw[line width=0.15mm](\c+\t,0,\d) -- (\c+\t,0,\a+\d);
			\draw[line width=0.15mm](\c+\t,\a,\d) -- (\c+\t,\a,\a+\d);
			\draw[line width=0.15mm](\t,\a,\d) -- (\t,\a,\a+\d);

			\filldraw[#5] (\t+\b,\b,\a+\d) -- (\c+\t-\b,\b,\a+\d) -- (\c+\t-\b,\a-\b,\a+\d) -- (\t+\b,\a-\b,\a+\d) -- (\t+\b,\b,\a+\d); 
			\filldraw[#5] (\t+\b,\a,\a-\b+\d) -- (\c+\t-\b,\a,\a-\b+\d) -- (\c+\t-\b,\a,\b+\d) -- (\t+\b,\a,\b+\d);
			\ifthenelse {\equal{#5} {}}
			{} 
			{\filldraw[#5] (\c+\t,\b,\a-\b+\d) -- (\c+\t,\b,\b+\d) -- (\c+\t,\a-\b,\b+\d) -- (\c+\t,\a-\b,\a-\b+\d);} 
		}
		
\def\connectvert#1#2#3{%
    \path let
      \p1 = ($(#2)-(#1)$),
      \n1 = {veclen(\p1)-0.cm},
      \n2 = {atan2(\x1,\y1)} 
    in
      (#1) -- (#2) node[#3, midway, sloped, rotate = 0, shading angle =  \n2+180, minimum height=\n1, minimum width=2pt, inner sep=3pt] {};
            }

\usepackage[linesnumbered,ruled,vlined]{algorithm2e}

\SetCommentSty{mycommfont}
\SetKwInput{Input}{Input}                
\SetKwInput{Output}{Output}         
\SetKwInput{Dataset}{Dataset}     
\SetKwInput{Parameters}{Parameters}  
\SetKwInput{Settings}{Settings}      
\SetKwInput{Initialization}{Initialization}      
\SetAlgorithmName{Alg.}{Alg.}{List of Algorithms} 
\makeatletter
\newcommand{\algorithmfootnote}[2][\footnotesize]{%
  \let\old@algocf@finish\@algocf@finish
  \def\@algocf@finish{\old@algocf@finish
    \leavevmode\rlap{\begin{minipage}{\linewidth}
    #1#2
    \end{minipage}}%
  }%
}
\makeatother

\usepackage{slashbox}

\usepackage{floatrow}
\usepackage{caption} 
\usepackage{makecell}

\usepackage{booktabs}
\usepackage{multirow}
\newcolumntype{P}[1]{>{\centering\arraybackslash}p{#1}}

\usepackage{hyperref} 
  
\newtheoremstyle{bfnote}%
  {}{}
  {\itshape}{}
  {\bfseries}{.}
  { }{\thmname{#1}\thmnumber{ #2}\thmnote{ (#3)}}
\theoremstyle{bfnote}

\usepackage{url}

\newcommand\blfootnote[1]{%
  \begingroup
  \renewcommand\thefootnote{}\footnote{#1}%
  \addtocounter{footnote}{-1}%
  \endgroup
}

\pgfplotsset{compat=1.18}

%% file: sec/abs.tex
\begin{abstract} 
Recent learning-based approaches have made astonishing advances in calibrated medical imaging like computerized tomography (CT), yet they struggle to generalize in uncalibrated modalities -- notably magnetic resonance (MR) imaging, where performance is highly sensitive to the differences in MR contrast, resolution, and orientation. This prevents broad applicability to diverse real-world clinical protocols. Here we introduce \texttt{BrainFM}, a modality-agnostic, multi-task vision foundation model for human brain imaging. With the proposed ``mild-to-severe'' intra-subject generation and ``real-synth'' mix-up training strategy, \texttt{BrainFM} is resilient to the appearance of acquired images (e.g., modality, contrast, deformation, resolution, artifacts), and can be directly applied to five fundamental brain imaging tasks, including image synthesis for CT and T1w/T2w/FLAIR MRI, anatomy segmentation, scalp-to-cortical distance, bias field estimation, and registration. We evaluate the efficacy of \texttt{BrainFM} on eleven public datasets, and demonstrate its robustness and effectiveness across all tasks and input modalities. Code is available at \href{https://github.com/jhuldr/BrainFM}{https://github.com/jhuldr/BrainFM}. 
\end{abstract}

\keywords{
Human Brain Imaging, Modality-agnostic Learning, Foundation Model
}

%% file: sec/intro.tex
 
\section{Introduction}
\label{sec: intro}  

Magnetic resonance imaging (MRI) enables in vivo noninvasive imaging of the human brain with exquisite and tunable soft-tissue contrast~\cite{BrantZawadzki1992MPRA}. Recent machine learning based methods have achieved great improvements in faster and more accurate image analysis of brain MRI, such as image segmentation~\cite{Ronneberger2015UNetCN, Milletar2016VNetFC, Kamnitsas2016EfficientM3, Ding_2021_ICCV,Iglesias2023SynthSRAP}, registration~\cite{Balakrishnan2018VoxelMorphAL,Yang2017QuicksilverFP,de2019deep,Shen2021Accurate}, super-resolution~\cite{Tian2020ImprovingIV,Tanno2020UncertaintyMI}, and connectivity studies~\cite{Mller2011UnderconnectedBH}. However, most existing MRI analysis methods are specific to certain MR contrast(s) and often require near-isotropic acquisitions. Therefore, models face sharp performance drops when voxel size and anisotropy increase, or are being used for a different contrast than what seen during training~\cite{wang2018deep}. This reduces model generalizability and results in duplicate data collection and training efforts given new datasets. By resorting to synthetic data, recent contrast-agnostic models~\cite{Iglesias2020JointSA,Liu2021YETI,Liu2022SONATA,Iglesias2023SynthSRAP,Billot2021SynthSegSO,Hoffmann2020SynthMorphLC,Laso2023WMH,gopinath2025recon} achieve impressive results and largely extend the applicability of models to heterogeneous clinical acquisition protocols. However, these models are only applicable to the tasks they were trained for.

Meanwhile, task-agnostic foundation models in computer vision and natural language processing have witnessed remarkable success, fueled by fast developments in computing and  large-scale datasets~\cite{Brown2020LanguageMA, Chowdhery2022PaLMSL, Kirillov2023SegmentA}. Often designed in a task-agnostic manner, foundation models have shown impressive performance in obtaining  general feature representations that can be quickly adapted (e.g., finetuned) to a wide range of downstream tasks~\cite{Bommasani2021OnTO, Awais2023FoundationalMD}. However, due to different acquisition protocols, processing pipelines, and privacy requirements across institutions, assembling large-scale datasets in medical imaging requires significantly more effort than in natural language processing or computer vision. As a result, medical foundation models are not as well developed. The MONAI~\cite{cardoso2022monai} project includes a model zoo with pretrained models for various tasks, yet they are all highly task-oriented and sensitive to specific image contrasts. Zhou et al.~\cite{Zhou2023AFM} constructed a medical foundation model, yet it is designed for the detection of eye and systemic health conditions from retinal scans, and only works on the modalities of color fundus photography and optical coherence tomography. More recently, generalist biomedical AI systems trained with vision-language data~\cite{Moor2023FoundationMF,Singhal2022LargeLM, Tu2023TowardsGB} have shown great potential in biomedical tasks, e.g., (visual) question answering, image classification, or radiology report generation and summarization. However, they have not explored  challenging pixel-wise  vision tasks such as reconstruction, segmentation, super-resolution, and registration.

In this work, we introduce \texttt{BrainFM}, a modality-agnostic, multi-task vision foundation model for human brain imaging.
\begin{itemize}
    \item[1)] Our on-the-fly, intra-subject data generator that can synthesize \textit{any} contrast and resolution, with a progressive mild-to-severe scheme to randomly corrupt images during training~(\cref{fig: augment}, \cref{sec: sample}). A ``real-synth'' mix-up strategy is further adopted to train \texttt{BrainFM} in a much more expansive and diverse space.
    
    \item[2)] \texttt{BrainFM} is trained in a multi-task fashion, covering five fundamental human brain imaging tasks including:
    joint synthesis and super-resolution of four modalities (T1-weighted MRI, T2-weighted MRI, FLAIR MRI,  and CT), anatomy segmentation, prediction of cortical distance maps, atlas registration, and bias field estimation.
    
    \item[3)] We extensively evaluate \texttt{BrainFM} across all five tasks on eleven public datasets (over 5,000 images in total), including input modalities of MR (T1w, T2w, FLAIR) and CT. We further adapt \texttt{BrainFM} to hemisphere and low-field image domains, and to new downstream tasks of image super-resolution and brain age prediction. \texttt{BrainFM} achieves state-of-the-art performance in all tasks and input image modalities.
    
\end{itemize}

This manuscript is a significant extension of our previous conference publication, where we presented our \texttt{Brain-ID} model~\cite{Liu2024BrainID}. Specifically,  we extend \texttt{Brain-ID} to \texttt{BrainFM} into a 
multi-task foundation model, and provide more extensive experimental evaluations from multiple aspects. (1)~We introduce an improved anatomy segmentation labeling system and a ``real-synth'' mix-up strategy for more realistic training data generation; (2)~We extend \texttt{Brain-ID} to five brain imaging tasks during pre-training, making it ready to use without additional fine-tuning effort.

%% file: sec/related_work.tex
\section{Related work}
\label{sec: related_work}
\subsection{Foundation Models in Medical Imaging}
\label{related: feat_repre} 
As discussed in Sec.~\ref{sec: intro}, general feature representation learning in the medical imaging can be more challenging than in the natural image domain, due to limited data availability. Xu et al.~\cite{Xu2014DeepLO} introduced a multiple instance learning model for feature representation in medical imaging, but it is designed specifically for classification, and only applies to histopathology images. You et al.~\cite{You2021MomentumCV} presented CVRL, a semi-supervised approach for voxel-wise representations, which is designed for image segmentation, but requires CT inputs to extract anatomical information. SAM~\cite{Yan2020SAMSL} is a self-supervised framework to encode anatomical information from CT images for feature embeddings, which has shown to be effective in downstream tasks such as registration (SAMConvex~\cite{Li2023SAMConvexFD}). However, same as CVRL, SAM only works on CT. BrainPrint~\cite{Wachinger2014BrainPrintI} is a compact and discriminative representation of brain morphology, which is specifically designed for cortical surface analyses. To the best of our knowledge, CIFL~\cite{chua2023contrast} is the only existing work on learning contrast-agnostic and task-independent brain feature representations. CIFL relies on contrastive learning alone, and is insufficient to outperform task/contrast-specific supervised models in downstream applications~\cite{Liu2024BrainID}. 

\subsection{Contrast-agnostic Learning for MR Images}
\label{related: contrast_agnostic} 
MRI scans acquired at different sites sites vary substantially in MRI contrast, resolution, and orientation. When given a new dataset, heterogeneity leads to duplicate training efforts for approaches that are sensitive to specific combinations of MR contrast, resolution, or orientation. Many classical brain segmentation models used Bayesian inference for contrast-robustness~\cite{Leemput2003AUF, Fischl2002WholeBS}, which requires a long processing time and struggles with low or anisotropic resolutions~\cite{Puonti2016FastAS,Iglesias2023SynthSRAP}. Recently, SynthSeg~\cite{Billot2021SynthSegSO,Laso2023WMH} was proposed for contrast-agnostic segmentation and achieves impressive results with a synthetic generator that simulates widely diverse contrasts. Meanwhile, there have been works with similar ideas of using synthetic data to achieve contrast-invariance in tasks like image registration~\cite{Hoffmann2020SynthMorphLC}, super-resolution~\cite{Iglesias2020JointSA}, or  skullstripping~\cite{Hoopes2022SynthStripSF}. However, all the above-mentioned methods are trained in a task-specific manner, whose features therefore cannot be readily applied to other domains. 

%% file: sec/method/main.tex
\input{sec/method/fig/fw_gen}


\section{Method}
\label{sec: method}

As mentioned in \cref{sec: intro}, the main challenges to obtain a general and robust foundation model for medical imaging lie in \textit{(i)}~the practical restrictions of building large-scale datasets with diverse modalities and contrasts; and \textit{(ii)}~the nature of most medical imaging models that are task-oriented and specific to data type (modality, contrast, resolution, orientation, etc). We aim to build a general brain imaging foundation model that is:

\noindent \textit{(i)} \textit{Robust}: the model should be robust to each subject's distinct anatomy, unaffected by variations in image modality, contrast, poses/deformations, resolutions, or artifacts.

\noindent \textit{(ii)} \textit{Expressive}: the well-trained model should also exhibit high expressiveness, containing rich information that facilitates easy and effective adaptation to diverse downstream tasks, eliminating the necessity for extensive training data and re-training efforts.

We first introduce \texttt{BrainFM}'s data generator (\cref{sec: generator}) that synthesizes a diverse range of data. Then we present our training pipeline (\cref{sec: framework}) for modality-agnostic multi-task learning.


\input{sec/method/generator}

\input{sec/method/fig/fw_train}

\input{sec/method/trainer}

\input{sec/method/fig/fw_multitask}



%% file: sec/method/fig/fw_gen.tex
 
\begin{figure*}[t]
\centering
\resizebox{\textwidth}{!}{
	
	\begin{tikzpicture}[lattice/.cd,spacing/.initial=4,superlattice
  period/.initial=12,amplitude/.initial=2]
	\pgfmathsetmacro{\cubex}{0.29*3}
	\pgfmathsetmacro{\cubey}{0.29*3}
	\pgfmathsetmacro{\cubez}{0.028*3}
	

	\pgfmathsetmacro{\shift}{0.3}
	\foreach \i in {-1.15}
	{
	\draw[black,fill=gray!30, line width = 0.02mm] (\i+\shift+0.1*3,0.15*3,0.1*3) -- ++(-\cubex,0,0) -- ++(0,-\cubey,0) -- ++(\cubex,0,0) -- cycle;
	\draw[black,fill=gray!35, line width = 0.02mm] (\i+\shift+0.1*3,0.15*3,0.1*3) -- ++(0,0,-\cubez) -- ++(0,-\cubey,0) -- ++(0,0,\cubez) -- cycle;
	\draw[black,fill=gray!35, line width = 0.02mm] (\i+\shift+0.1*3,0.15*3,0.1*3) -- ++(-\cubex,0,0) -- ++(0,0,-\cubez) -- ++(\cubex,0,0) -- cycle;
	}
 
	\node at (-1.15+\shift+0.2975*3, 0.3475*3, 0.99*3) {\includegraphics[width=0.048\textwidth]{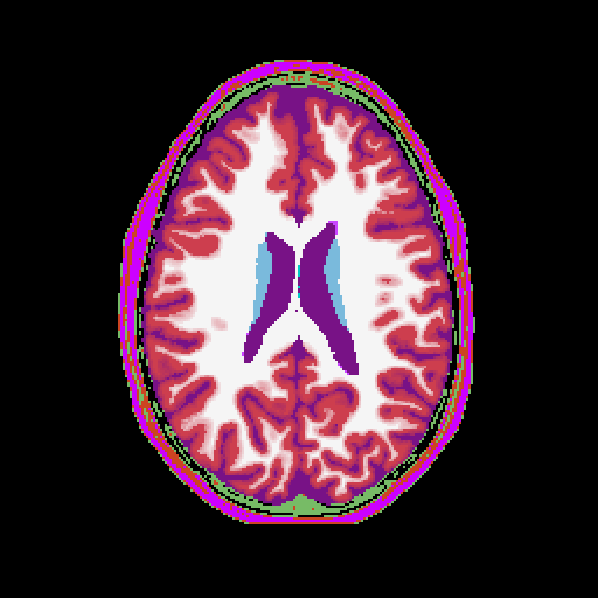}};

	\draw[-latex] (-1.17+0.6+\shift, 0.05, 0.25*3) -- (0.1+2., 0.05, 0.25*3);
	\node at (-1.15+0.93+\shift, 0.49, 3.1) {\tiny Label ($L$)}; 
	\node at (-2.01+1.8+\shift, 0.2, 0.25*3) {\tiny \cref{eq: deform}};
	\node at (-1.0+2.1+\shift, 0.2, 0.25*3) {\tiny \text{Deformation}};
 
	\node at (-2+1.8+\shift, -0.1, 0.25*3) {\tiny \cref{eq: contrast}}; 
	\node at (-1.13+2.04+\shift, -0.1, 0.25*3) {\tiny \text{Contrast}};

	\pgfmathsetmacro{\dx}{-5.5}
	\pgfmathsetmacro{\dy}{2.7}
        
        \begin{scope}[xshift=-5]
            \pgftransformnonlinear{\latticetilt}
            \draw[orange!80,step=0.08,thin] (0.15,0.15) grid (0.88,0.88);
            \draw[->, black] (0.15,0.15) -- +(0.85,0);
            \draw[->, black] (0.15,0.15) -- +(0,0.85);
            
        \end{scope} 
    
	\draw[->, line width = 0.2mm, color=orange!80] (5.5+\dx+\shift, -2.6+\dy) -- (5.5+\dx+\shift, -2.9+\dy);
 
	\node at (0.35+\shift+0.2975*3, 0.26, 3.2) {\includegraphics[width=0.074\textwidth]{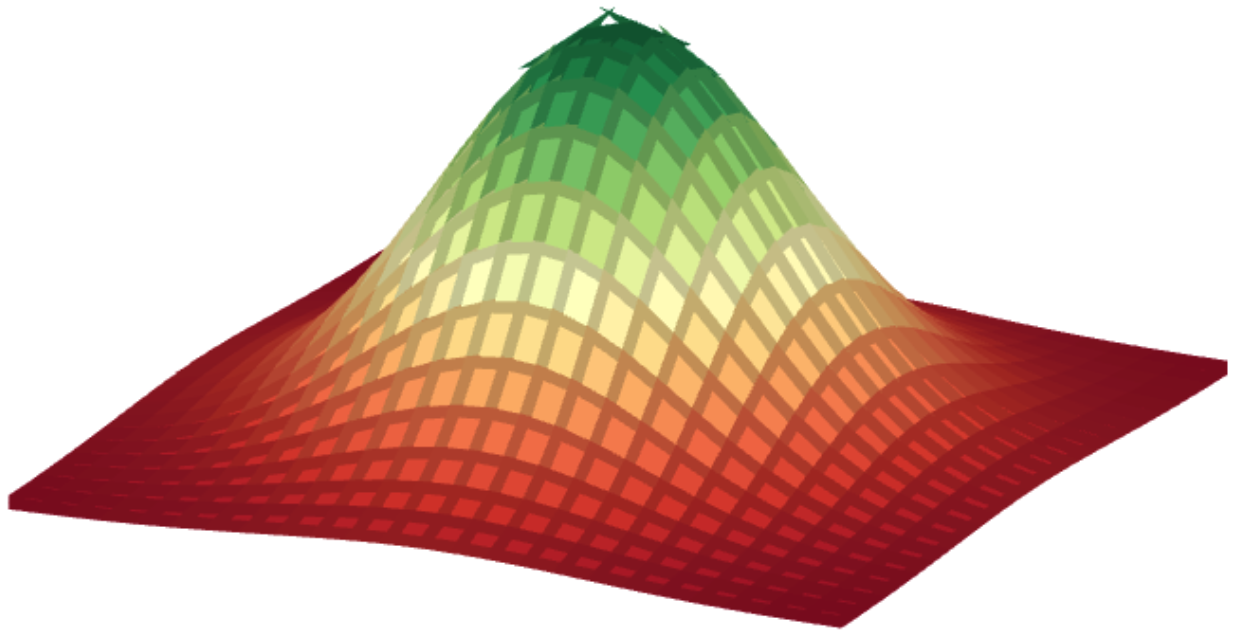}};

	\draw[->, line width = 0.2mm, color=matcha!200] (5.5+\dx+\shift, -3.28+\dy) -- (5.5+\dx+\shift, -2.98+\dy);
	
	
	\pgfmathsetmacro{\shift}{1.3}
	\foreach \i in {1.3}
	{
	\draw[black,fill=gray!30, line width = 0.02mm] (\i+\shift+0.4*3,0.5*3,0.5*3) -- ++(-\cubex,0,0) -- ++(0,-\cubey,0) -- ++(\cubex,0,0) -- cycle;
	\draw[black,fill=gray!35, line width = 0.02mm] (\i+\shift+0.4*3,0.5*3,0.5*3) -- ++(0,0,-\cubez) -- ++(0,-\cubey,0) -- ++(0,0,\cubez) -- cycle;
	\draw[black,fill=gray!35, line width = 0.02mm] (\i+\shift+0.4*3,0.5*3,0.5*3) -- ++(-\cubex,0,0) -- ++(0,0,-\cubez) -- ++(\cubex,0,0) -- cycle;
	\node at (\i+\shift+0.409*3, 0.509*3, 0.9*3) {\includegraphics[width=0.048\textwidth]{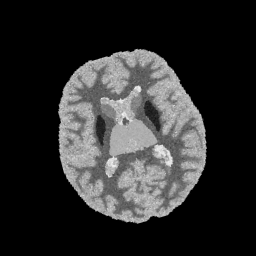}};
	\node at (\i+\shift-0.35, 0.26, 0.13) {\tiny $I_{i}$};

	\draw[black,fill=gray!30, line width = 0.02mm] (\i+\shift-0.1*3,-0.2*3,-0.2*3) -- ++(-\cubex,0,0) -- ++(0,-\cubey,0) -- ++(\cubex,0,0) -- cycle;
	\draw[black,fill=gray!35, line width = 0.02mm] (\i+\shift-0.1*3, -0.2*3,-0.2*3) -- ++(0,0,-\cubez) -- ++(0,-\cubey,0) -- ++(0,0,\cubez) -- cycle;
	\draw[black,fill=gray!35, line width = 0.02mm] (\i+\shift-0.1*3,-0.2*3,-0.2*3) -- ++(-\cubex,0,0) -- ++(0,0,-\cubez) -- ++(\cubex,0,0) -- cycle;	
	\node at (\i+\shift-0.75, -0.35*3, -0.212*3) {\includegraphics[width=0.048\textwidth]{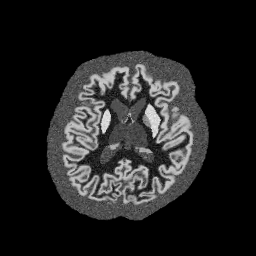}}; 
 
	\draw[black,fill=black] (\i+\shift+0.45, 0., 0.2*3) circle (0.4pt);
	\draw[black,fill=black] (\i+\shift+0.45, 0., 0.25*3) circle (0.4pt);
	\draw[black,fill=black] (\i+\shift+0.45, 0., 0.3*3) circle (0.4pt);
	\node at (\i+\shift+0.1, 0.08, 3.1) {\tiny $I_{1}$}; 
	
	\draw[-latex] (\i+\shift+0.7, 0.05, 0.25*3) -- (\i+\shift+1.93, 0.05, 0.25*3); 
	\node at (\i+\shift+1.30, 0.2, 0.25*3) {\tiny \text{Corruption}};
	\node at (\i+\shift+1.20, -0.1, 0.25*3) {\tiny \text{Resampling}};
	\node at (\i+\shift+1.19, -0.3, 0.25*3) {\tiny \text{({Resolution}}};
	\node at (\i+\shift+1.21, -0.45, 0.25*3) {\tiny \text{simulation)}};
	
	}
	
	
	\pgfmathsetmacro{\cubeza}{0.05*3}
	\pgfmathsetmacro{\cubezb}{0.04*3}
	\pgfmathsetmacro{\shift}{4.2}
	\foreach \i in {1}
	{
	\draw[black,fill=gray!30, line width = 0.02mm] (\i+\shift+0.4*3,0.5*3,0.5*3) -- ++(-\cubex,0,0) -- ++(0,-\cubey,0) -- ++(\cubex,0,0) -- cycle;
	\draw[black,fill=gray!35, line width = 0.02mm] (\i+\shift+0.4*3,0.5*3,0.5*3) -- ++(0,0,-\cubeza) -- ++(0,-\cubey,0) -- ++(0,0,\cubeza) -- cycle;
	\draw[black,fill=gray!35, line width = 0.02mm] (\i+\shift+0.4*3,0.5*3,0.5*3) -- ++(-\cubex,0,0) -- ++(0,0,-\cubeza) -- ++(\cubex,0,0) -- cycle;
	\node at (\i+\shift+0.4086*3, 0.509*3, 0.9*3) {\includegraphics[width=0.048\textwidth]{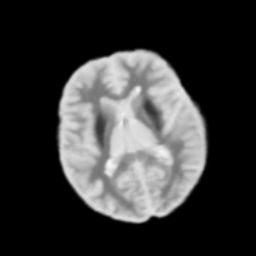}};
	\node at (\i+\shift+0.2, -0.05, 0.13) {\tiny $S_{i-\i}$};

	\draw[black,fill=gray!30, line width = 0.02mm] (\i+\shift-0.1*3,-0.2*3,-0.2*3) -- ++(-\cubex,0,0) -- ++(0,-\cubey,0) -- ++(\cubex,0,0) -- cycle;
	\draw[black,fill=gray!35, line width = 0.02mm] (\i+\shift-0.1*3, -0.2*3,-0.2*3) -- ++(0,0,-\cubezb) -- ++(0,-\cubey,0) -- ++(0,0,\cubezb) -- cycle;
	\draw[black,fill=gray!35, line width = 0.02mm] (\i+\shift-0.1*3,-0.2*3,-0.2*3) -- ++(-\cubex,0,0) -- ++(0,0,-\cubezb) -- ++(\cubex,0,0) -- cycle;
	
	\node at (\i+\shift-0.241*3, -0.341*3, -0.19*3) {\includegraphics[width=0.048\textwidth]{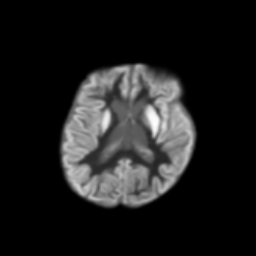}};
	\node at (\i+\shift+0.65, -0.23, 3.1) {\tiny $S_{1-\i}$};
	}

	\pgfmathsetmacro{\cubeza}{0.03*3}
	\pgfmathsetmacro{\cubezb}{0.1*3}
	\pgfmathsetmacro{\shift}{4.2}
	\foreach \i in {2}
	{
	\draw[black,fill=gray!30, line width = 0.02mm] (\i+\shift+0.4*3,0.5*3,0.5*3) -- ++(-\cubex,0,0) -- ++(0,-\cubey,0) -- ++(\cubex,0,0) -- cycle;
	\draw[black,fill=gray!35, line width = 0.02mm] (\i+\shift+0.4*3,0.5*3,0.5*3) -- ++(0,0,-\cubeza) -- ++(0,-\cubey,0) -- ++(0,0,\cubeza) -- cycle;
	\draw[black,fill=gray!35, line width = 0.02mm] (\i+\shift+0.4*3,0.5*3,0.5*3) -- ++(-\cubex,0,0) -- ++(0,0,-\cubeza) -- ++(\cubex,0,0) -- cycle;
	\node at (\i+\shift+0.4086*3, 0.509*3, 0.9*3) {\includegraphics[width=0.048\textwidth]{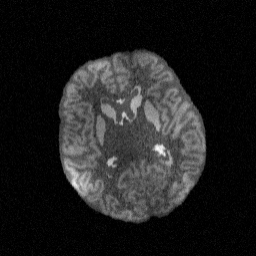}};
	\node at (\i+\shift+0.2, -0.05, 0.13) {\tiny $S_{i-\i}$};

	\draw[black,fill=gray!30, line width = 0.02mm] (\i+\shift-0.1*3,-0.2*3,-0.2*3) -- ++(-\cubex,0,0) -- ++(0,-\cubey,0) -- ++(\cubex,0,0) -- cycle;
	\draw[black,fill=gray!35, line width = 0.02mm] (\i+\shift-0.1*3, -0.2*3,-0.2*3) -- ++(0,0,-\cubezb) -- ++(0,-\cubey,0) -- ++(0,0,\cubezb) -- cycle;
	\draw[black,fill=gray!35, line width = 0.02mm] (\i+\shift-0.1*3,-0.2*3,-0.2*3) -- ++(-\cubex,0,0) -- ++(0,0,-\cubezb) -- ++(\cubex,0,0) -- cycle;
	
	\node at (\i+\shift-0.241*3, -0.341*3, -0.19*3) {\includegraphics[width=0.048\textwidth]{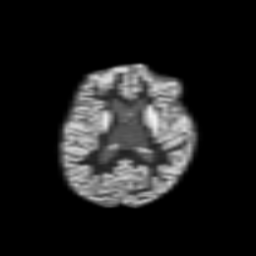}};
	\node at (\i+\shift+0.65, -0.23, 3.1) {\tiny $S_{1-\i}$};
	}

	\pgfmathsetmacro{\cubeza}{0.05*3}
	\pgfmathsetmacro{\cubezb}{0.04*3}
	\pgfmathsetmacro{\shift}{4.2}
	\foreach \i in {3}
	{
	\draw[black,fill=gray!30, line width = 0.02mm] (\i+\shift+0.4*3,0.5*3,0.5*3) -- ++(-\cubex,0,0) -- ++(0,-\cubey,0) -- ++(\cubex,0,0) -- cycle;
	\draw[black,fill=gray!35, line width = 0.02mm] (\i+\shift+0.4*3,0.5*3,0.5*3) -- ++(0,0,-\cubeza) -- ++(0,-\cubey,0) -- ++(0,0,\cubeza) -- cycle;
	\draw[black,fill=gray!35, line width = 0.02mm] (\i+\shift+0.4*3,0.5*3,0.5*3) -- ++(-\cubex,0,0) -- ++(0,0,-\cubeza) -- ++(\cubex,0,0) -- cycle;
	\node at (\i+\shift+0.4086*3, 0.509*3, 0.9*3) {\includegraphics[width=0.048\textwidth]{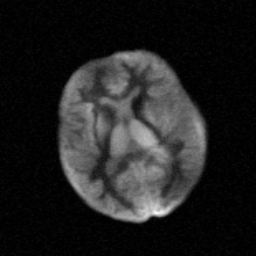}};
	\node at (\i+\shift+0.2, -0.05, 0.13) {\tiny $S_{i-\i}$};
	\draw[black,fill=black] (\i+\shift+1.1, 0.4, 0.25*3) circle (0.4pt);
	\draw[black,fill=black] (\i+\shift+1.2, 0.4, 0.25*3) circle (0.4pt);
	\draw[black,fill=black] (\i+\shift+1.3, 0.4, 0.25*3) circle (0.4pt);

	\draw[black,fill=gray!30, line width = 0.02mm] (\i+\shift-0.1*3,-0.2*3,-0.2*3) -- ++(-\cubex,0,0) -- ++(0,-\cubey,0) -- ++(\cubex,0,0) -- cycle;
	\draw[black,fill=gray!35, line width = 0.02mm] (\i+\shift-0.1*3, -0.2*3,-0.2*3) -- ++(0,0,-\cubezb) -- ++(0,-\cubey,0) -- ++(0,0,\cubezb) -- cycle;
	\draw[black,fill=gray!35, line width = 0.02mm] (\i+\shift-0.1*3,-0.2*3,-0.2*3) -- ++(-\cubex,0,0) -- ++(0,0,-\cubezb) -- ++(\cubex,0,0) -- cycle;
	
	\node at (\i+\shift-0.241*3, -0.341*3, -0.19*3) {\includegraphics[width=0.048\textwidth]{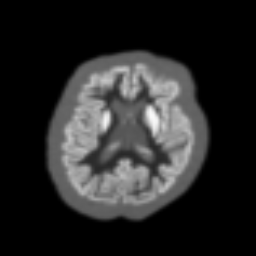}};
	\node at (\i+\shift+0.65, -0.23, 3.1) {\tiny $S_{1-\i}$};
	}

	\pgfmathsetmacro{\cubeza}{0.07*3}
	\pgfmathsetmacro{\cubezb}{0.02*3}
	\pgfmathsetmacro{\shift}{4.2}
	\foreach \i in {4}
	{
	\draw[black,fill=gray!30, line width = 0.02mm] (\i+\shift+0.4*3,0.5*3,0.5*3) -- ++(-\cubex,0,0) -- ++(0,-\cubey,0) -- ++(\cubex,0,0) -- cycle;
	\draw[black,fill=gray!35, line width = 0.02mm] (\i+\shift+0.4*3,0.5*3,0.5*3) -- ++(0,0,-\cubeza) -- ++(0,-\cubey,0) -- ++(0,0,\cubeza) -- cycle;
	\draw[black,fill=gray!35, line width = 0.02mm] (\i+\shift+0.4*3,0.5*3,0.5*3) -- ++(-\cubex,0,0) -- ++(0,0,-\cubeza) -- ++(\cubex,0,0) -- cycle;
	\node at (\i+\shift+0.4086*3, 0.509*3, 0.9*3) {\includegraphics[width=0.048\textwidth]{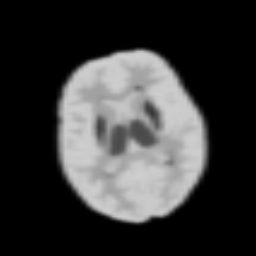}};
	\node at (\i+\shift+0.2, -0.05, 0.13) {\tiny $S_{i-\i}$};
	\draw[black,fill=black] (\i+\shift+1.1, 0.4, 0.25*3) circle (0.4pt);
	\draw[black,fill=black] (\i+\shift+1.2, 0.4, 0.25*3) circle (0.4pt);
	\draw[black,fill=black] (\i+\shift+1.3, 0.4, 0.25*3) circle (0.4pt);

	\draw[black,fill=gray!30, line width = 0.02mm] (\i+\shift-0.1*3,-0.2*3,-0.2*3) -- ++(-\cubex,0,0) -- ++(0,-\cubey,0) -- ++(\cubex,0,0) -- cycle;
	\draw[black,fill=gray!35, line width = 0.02mm] (\i+\shift-0.1*3, -0.2*3,-0.2*3) -- ++(0,0,-\cubezb) -- ++(0,-\cubey,0) -- ++(0,0,\cubezb) -- cycle;
	\draw[black,fill=gray!35, line width = 0.02mm] (\i+\shift-0.1*3,-0.2*3,-0.2*3) -- ++(-\cubex,0,0) -- ++(0,0,-\cubezb) -- ++(\cubex,0,0) -- cycle;
	
	\node at (\i+\shift-0.241*3, -0.341*3, -0.19*3) {\includegraphics[width=0.048\textwidth]{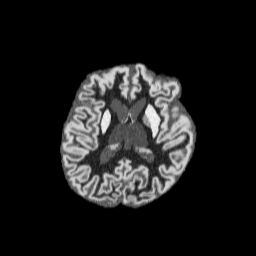}};
	\node at (\i+\shift+0.65, -0.23, 3.1) {\tiny $S_{1-\i}$};
	\draw[black,fill=black] (\i+\shift+0.36, -0.93, 0.25*3) circle (0.4pt);
	\draw[black,fill=black] (\i+\shift+0.46, -0.93, 0.25*3) circle (0.4pt);
	\draw[black,fill=black] (\i+\shift+0.56, -0.93, 0.25*3) circle (0.4pt);
	}
	\end{tikzpicture}
	
} 
\caption{\texttt{BrainFM}'s data generator on the fly. Given the brain segmentation labels of a subject, we randomly generate a deformation field, and synthesize intra-subjecct samples featuring various contrast intensities and corruption levels (\cref{sec: generator}).}
\label{fig: augment}
\end{figure*}
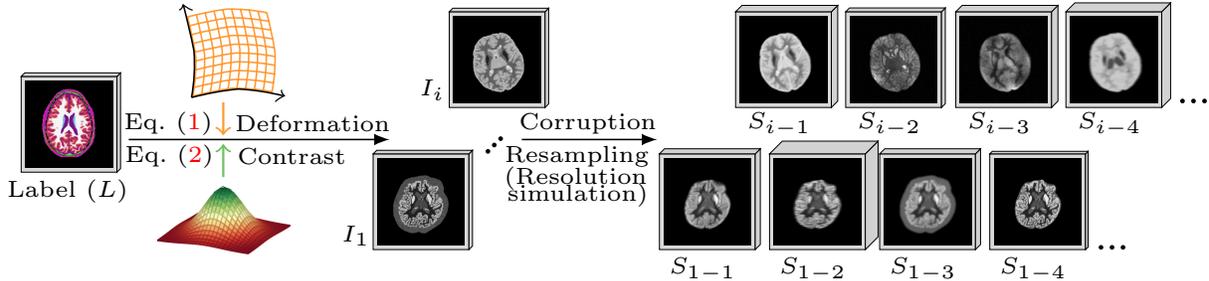

%% file: sec/method/generator.tex
\subsection{Enriching the Intra-subject Learning Space}
\label{sec: generator}

A robust representation relies on large-scale data, however, different acquisition protocols, processing pipelines, and privacy requirements across institutions make large-scale data less accessible and require significantly more effort, e.g., via federated learning~\cite{Sheller2020Federated,Adnan2022Federated,Darzidehkalani2022Federated}. Moreover, only limited acquired images/modalities are usually available per subject. This lack of data consistency is a significant barrier to obtaining a robust, modality-agnostic model. \texttt{BrainFM} avoids these barriers by generating synthetic data to enrich the learning space. 

To generate images with complex brain structures, we start from high-resolution brain segmentation images that provide labels of brain structures and extracerebral regions ($L$ in \cref{fig: augment}). The generatation consists of three steps, \textit{(i)}~deformation generation, \textit{(ii)}~contrast synthesis, and \textit{(iii)}~data corruption (including lower resolution resampling). For simplicity, $\Theta$ denotes the parameter group of the generation process described below.


\subsubsection{Generating deformations}
\label{sec: deform}
We first generate a random deformation field ($\phi\vert_{\theta_\phi}$) consisting of an affine transformation and a non-linear displacement field:
\begin{equation}
    \phi\vert_{\theta_\phi} = \mathcal{T}\vert_{\theta_\phi} \circ \mathcal{A}\vert_{\theta_\phi}\,,
    \label{eq: deform}
\end{equation}
where $\mathcal{A}\vert_{\theta_\phi}$ denotes an affine transformation matrix which includes affine (rotation, scaling, shearing, tranlation) transformation, and $\mathcal{T}\vert_{\theta_\phi}$ refers to a non-linear displacement field computed as the integration of a stationary velocity field (SVF) that is smooth and invertible everywhere and thus preserves the topology of the brain anatomy. $\theta_\phi \in \Theta$ controls the transformation ranges. Further details can be found in~\cite{Iglesias2020JointSA}.


\subsubsection{Simulation of random contrast}
\label{sec: contrast}
Then, we synthesize images ($I(x),\, x \in \Omega$) by randomly ``painting'' intensities on the segmentation maps according to their brain structure labels ($l\in L$). Specifically, the regional intensities are generated by separately sampling a Gaussian distribution on each labeled region:
\begin{equation} 
    \begin{cases}
        I(x) \sim \mathcal{N}(\mu_l,\, \sigma_l ) \,, \quad l \in L\,,\\ 
        \mu_l \sim \mathcal{N}(0,\, 1 \, \vert\, \theta_{\mu}, \, \theta_{l}) \,,~
        \sigma_l \sim \mathcal{N}(0,\, 1 \, \vert \, \theta_{\sigma}, \, \theta_{l}) \,,
    \end{cases}
    \label{eq: contrast}
\end{equation}
where $\mu_l$ and $\sigma_l$ refer to the mean and standard deviation of each segmentation label $l$, and are independently sampled from Gaussian distributions at each voxel. The hyperparameters $\theta_{l},\, \theta_{\mu},\, \theta_{\sigma} \in \Theta$ control the distribution of their random values. 


\subsubsection{Resolution Simulation and Data Corruption}
\label{sec: corrupt}
Given a deformed, contrast-synthesized image ($I$), we adopt the 
data corruption pipeline from~\cite{Iglesias2023SynthSRAP}, which further augments images with different levels of resolution and scanning artifacts that are commonly found in real-world clinical protocols.


\subsubsection{``Real-synth'' Data Mix-up}
\label{sec: mixup} 
During training, we integrate real and simulated data to enhance the realism of our simulations. Specifically, we randomly mix the intensities of simulated images with those from real images, if available, acquired from the same subject, to help bridge the gap between real and simulated imaging spaces and make the generated images better reflect the actual variations and complexities present in real-world data.

As illustrated in \cref{fig: augment}, \texttt{BrainFM} can generate an infinite number of variations from a single subject with its unique brain anatomy labels. By generating images with randomized contrast/resolution/orientation on the fly for each subject, we greatly enrich the learning space to ultimately obtain a robust representation. 

%% file: sec/method/fig/fw_train.tex
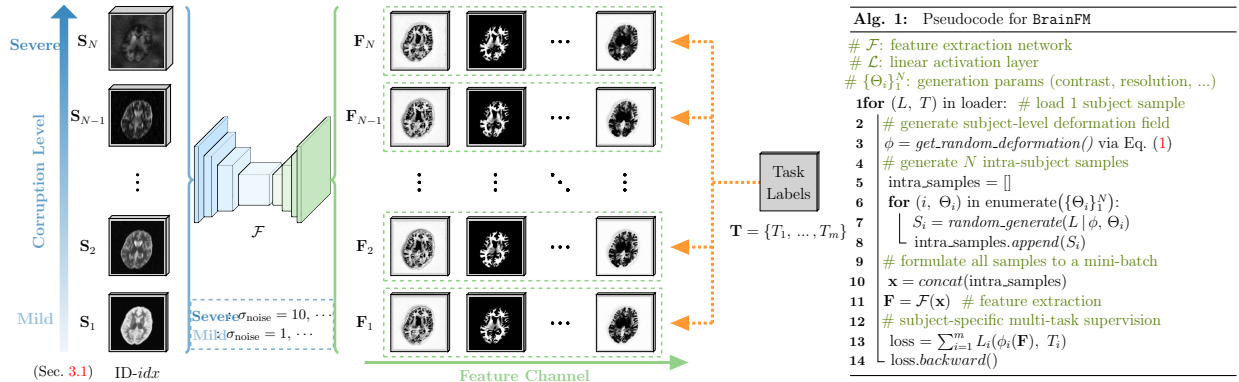
\begin{figure*}[t] 
\centering  
\resizebox{\linewidth}{!}{
	\begin{tikzpicture}
		\tikzstyle{myarrows}=[line width=0.8mm,draw=blue!50,-triangle 45,postaction={draw, line width=0.05mm, shorten >=0.02mm, -}]
		\tikzstyle{mylines}=[line width=0.8mm]
  


	\pgfmathsetmacro{\dx}{-1.5}
	\pgfmathsetmacro{\dy}{0}
        \connectvert{(-0+\dx, -7.13+\dy}{(-0+\dx, 2+\dy}{single arrow, top color=cb!10, bottom color=cb!150};
    \node at (0.85+\dx, -7.2+0.15*3, 0.75*3) {(\cref{sec: generator})};
    
	\node[rotate=90,anchor=north] at (-0.95+\dx, -2.6+\dy) {\textcolor{cb!120}{\large\textbf{Corruption Level}}};
 
	\node[] at (-0.75+\dx, -6.2+\dy) {\textcolor{cb!80}{\large\textbf{Mild}}};
	\node[] at (-0.75+\dx, 1.+\dy) {\textcolor{cb!150}{\large\textbf{Severe}}};


	\pgfmathsetmacro{\cubex}{0.5*3}
	\pgfmathsetmacro{\cubey}{0.5*3}

	\pgfmathsetmacro{\shift}{1.}
	\foreach \i/\cubez in {-5.9/0.12, -3.9/0.2, -0.5/0.3, 1.5/0.45}
	{
	\draw[black,fill=gray!35, line width = 0.02mm] (\shift+0.1*3,\i+0.15*3,0.1*3) -- ++(-\cubex,0,0) -- ++(0,-\cubey,0) -- ++(\cubex,0,0) -- cycle;
	\draw[black,fill=gray!35, line width = 0.02mm] (\shift+0.1*3,\i+0.15*3,0.1*3) -- ++(0,0,-\cubez) -- ++(0,-\cubey,0) -- ++(0,0,\cubez) -- cycle;
	\draw[black,fill=gray!35, line width = 0.02mm] (\shift+0.1*3,\i+0.15*3,0.1*3) -- ++(-\cubex,0,0) -- ++(0,0,-\cubez) -- ++(\cubex,0,0) -- cycle;
	}
 
	\node at (\shift+0.1*3, -5.9+0.15*3, 0.75*3) {\includegraphics[width=0.083\textwidth]{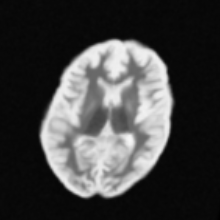}};
	\node at (\shift+0.1*3, -3.9+0.15*3, 0.75*3) {\includegraphics[width=0.083\textwidth]{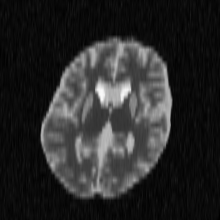}};
	\node at (\shift+0.1*3, -0.5+0.15*3, 0.75*3) {\includegraphics[width=0.083\textwidth]{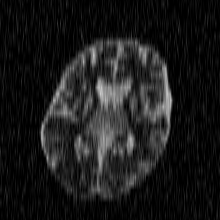}}; 
	\node at (\shift+0.1*3, 1.5+0.15*3, 0.75*3) {\includegraphics[width=0.083\textwidth]{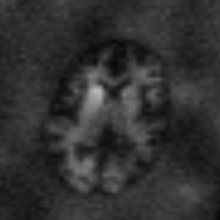}};
        \node at (0.+\shift+0.3-1.3, 1.5+0.15*3, 0.75*3) {\large$\mathbf{S}_N$};
        \node at (0.+\shift+0.3-1.3, -0.5+0.15*3, 0.75*3) {\large$\mathbf{S}_{N-1}$};
        \node at (0.+\shift+0.3-1.3, -3.9+0.15*3, 0.75*3) {\large$\mathbf{S}_2$};
        \node at (0.+\shift+0.3-1.3, -5.9+0.15*3, 0.75*3) {\large$\mathbf{S}_1$};
 
    \draw[black,fill=black] (\shift+0.3, -2.+0.15*3, 0.75*3) circle (0.8pt);
    \draw[black,fill=black] (\shift+0.3, -2.2+0.15*3, 0.75*3) circle (0.8pt);
    \draw[black,fill=black] (\shift+0.3, -2.4+0.15*3, 0.75*3) circle (0.8pt);

    \node at (\shift+0.3, -7.2+0.15*3, 0.75*3) {\large ID-$idx$};

    \node at (\shift+2.2, -6.2+0.15*3, 0.75*3) {\textcolor{cb!80}{\textbf{Mild}}};
    \node at (\shift+3.67, -6.2+0.15*3, 0.75*3) {: $\sigma_{\text{noise}} = 1,\, \cdots$};
    \node at (\shift+2.35, -5.7+0.15*3, 0.75*3) {\textcolor{cb!150}{\textbf{Severe}}};
    \node at (\shift+4.07, -5.7+0.15*3, 0.75*3) {: $\sigma_{\text{noise}} = 10,\, \cdots$};

        \pgfmathsetmacro{\dx}{-2.17}
        \pgfmathsetmacro{\dy}{-2.}
	\draw[dashed, color = cb, line width=0.4mm] (4+\dx, -5.+\dy) -- (7.75+\dx, -5.+\dy) -- (7.75+\dx, -3.7+\dy) -- (4+\dx, -3.7+\dy) -- (4+\dx, -5.+\dy);

    \draw [decorate,decoration={brace,amplitude=5pt,raise=6ex},line width=2.pt,color = cb] (0.7, 2) -- (0.7, -7.1);

    
    \pgfmathsetmacro{\sx}{2.1}
    \pgfmathsetmacro{\sy}{5.4}
    \pgfmathsetmacro{\dx}{\sx+0}
    \pgfmathsetmacro{\dy}{\sy+0.2} 
    \pgfmathsetmacro{\ddy}{-3.5} 
    \pgfmathsetmacro{\dxt}{\dx+0.5} 
    
    \node at (1.5+\sx, -6.+\dy+\ddy){\large $\mathcal{F}$};
    \pgfmathsetmacro{\dy}{\dy+\ddy} 
    \networkLayer{2}{0.1}{5.1-\dy+\dxt}{13-2.6*\dy}{color=myblue!80}{}
    \networkLayer{1.6}{0.2}{5.2-\dy+\dxt}{13-2.6*\dy}{color=myblue!60}{}
    \networkLayer{1.2}{0.4}{5.4-\dy+\dxt}{13-2.6*\dy}{color=myblue!40}{}
    \networkLayer{0.8}{0.8}{5.8-\dy+\dxt}{13-2.6*\dy}{color=myblue!20}{}
    
    \pgfmathsetmacro{\dxs}{\dx+0}
    \pgfmathsetmacro{\dys}{\dy+0} 
    \networkLayer{0.8}{0.8}{7.2-\dys+\dxs}{13-2.6*\dys}{color=matcha!20}{}
    \networkLayer{1.2}{0.4}{7.6-\dys+\dxs}{13-2.6*\dys}{color=matcha!40}{}
    \networkLayer{1.6}{0.2}{8.-\dys+\dxs}{13-2.6*\dys}{color=matcha!60}{}
    \networkLayer{2}{0.1}{8.3-\dys+\dxs}{13-2.6*\dys}{color=matcha!80}{}

 
    \draw [decorate,decoration={brace,amplitude=5pt,mirror,raise=6ex},line width=2.pt,color = matcha!150] (6.7, 2) -- (6.7, -7.1);

	\pgfmathsetmacro{\cubez}{0.12}  
	\pgfmathsetmacro{\shift}{1.2}
 
	\foreach \i/\j in {7.2/-5.9, 7.2/-3.9, 7.2/-0.5, 7.2/1.5,  9.2/-5.9, 9.2/-3.9, 9.2/-0.5, 9.2/1.5,  12.6/-5.9, 12.6/-3.9, 12.6/-0.5, 12.6/1.5}
	{
	\draw[black,fill=gray!35, line width = 0.02mm] (\i+\shift+0.1*3,\j+0.15*3,0.1*3) -- ++(-\cubex,0,0) -- ++(0,-\cubey,0) -- ++(\cubex,0,0) -- cycle;
	\draw[black,fill=gray!35, line width = 0.02mm] (\i+\shift+0.1*3,\j+0.15*3,0.1*3) -- ++(0,0,-\cubez) -- ++(0,-\cubey,0) -- ++(0,0,\cubez) -- cycle;
	\draw[black,fill=gray!35, line width = 0.02mm] (\i+\shift+0.1*3,\j+0.15*3,0.1*3) -- ++(-\cubex,0,0) -- ++(0,0,-\cubez) -- ++(\cubex,0,0) -- cycle;
 
        \draw[black,fill=black] (\i+\shift+0.3, -2.+0.15*3, 0.75*3) circle (0.8pt);
        \draw[black,fill=black] (\i+\shift+0.3, -2.2+0.15*3, 0.75*3) circle (0.8pt);
        \draw[black,fill=black] (\i+\shift+0.3, -2.4+0.15*3, 0.75*3) circle (0.8pt);
        
        \draw[black,fill=black] (10.9-0.2+\shift+0.3, \j+0.15*3, 0.75*3) circle (0.8pt);
        \draw[black,fill=black] (10.9+\shift+0.3, \j+0.15*3, 0.75*3) circle (0.8pt);
        \draw[black,fill=black] (10.9+0.2+\shift+0.3, \j+0.15*3, 0.75*3) circle (0.8pt);
        
        \draw[black,fill=black] (10.9-0.2+\shift+0.3, -2.+0.15*3, 0.75*3) circle (0.8pt);
        \draw[black,fill=black] (10.9+\shift+0.3, -2.2+0.15*3, 0.75*3) circle (0.8pt);
        \draw[black,fill=black] (10.9+0.2+\shift+0.3, -2.4+0.15*3, 0.75*3) circle (0.8pt);
	}

        \pgfmathsetmacro{\dx}{2.9}
        \foreach \dy in {-2.2, -0.2, 3.2, 5.2}
        {
	\draw[dashed, color = matcha!150, line width=0.4mm] (4+\dx, -5.+\dy) -- (11.3+\dx, -5.+\dy) -- (11.3+\dx, -3.2+\dy) -- (4+\dx, -3.2+\dy) -- (4+\dx, -5.+\dy); 
        }
        
	\node at (7.2+\shift+0.1*3, 1.5+0.15*3, 0.75*3) {\includegraphics[width=0.083\textwidth]{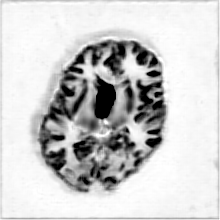}};
	\node at (7.2+\shift+0.1*3, -0.5+0.15*3, 0.75*3) {\includegraphics[width=0.083\textwidth]{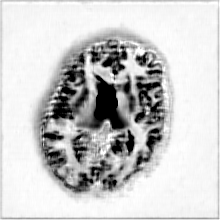}};
	\node at (7.2+\shift+0.1*3, -3.9+0.15*3, 0.75*3) {\includegraphics[width=0.083\textwidth]{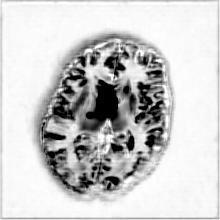}};
	\node at (7.2+\shift+0.1*3, -5.9+0.15*3, 0.75*3) {\includegraphics[width=0.083\textwidth]{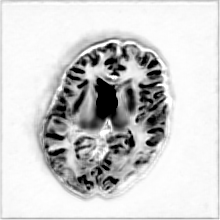}};
 
	\node at (9.2+\shift+0.1*3, 1.5+0.15*3, 0.75*3) {\includegraphics[width=0.083\textwidth]{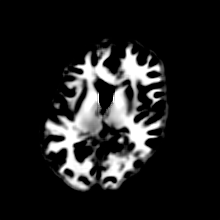}};
	\node at (9.2+\shift+0.1*3, -0.5+0.15*3, 0.75*3) {\includegraphics[width=0.083\textwidth]{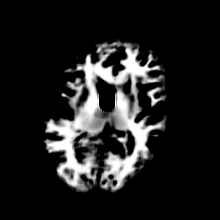}};
	\node at (9.2+\shift+0.1*3, -3.9+0.15*3, 0.75*3) {\includegraphics[width=0.083\textwidth]{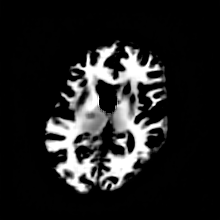}};
	\node at (9.2+\shift+0.1*3, -5.9+0.15*3, 0.75*3) {\includegraphics[width=0.083\textwidth]{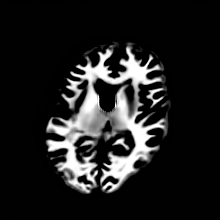}};
 
	\node at (12.6+\shift+0.1*3, 1.5+0.15*3, 0.75*3) {\includegraphics[width=0.083\textwidth]{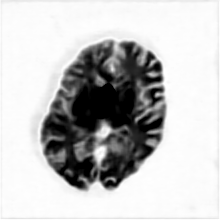}};
	\node at (12.6+\shift+0.1*3, -0.5+0.15*3, 0.75*3) {\includegraphics[width=0.083\textwidth]{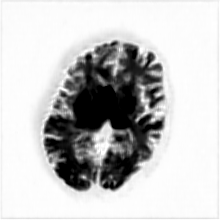}};
	\node at (12.6+\shift+0.1*3, -3.9+0.15*3, 0.75*3) {\includegraphics[width=0.083\textwidth]{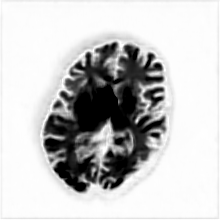}};
	\node at (12.6+\shift+0.1*3, -5.9+0.15*3, 0.75*3) {\includegraphics[width=0.083\textwidth]{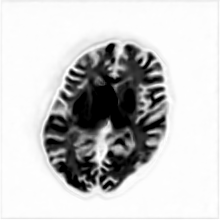}};

        \node at (7.27+\shift+0.3-1.5, 1.5+0.15*3, 0.75*3) {\large$\mathbf{F}_{N}$};
        \node at (7.25+\shift+0.3-1.5, -0.5+0.15*3, 0.75*3) {\large$\mathbf{F}_{N-1}$};
        \node at (7.27+\shift+0.3-1.5, -3.9+0.15*3, 0.75*3) {\large$\mathbf{F}_2$};
        \node at (7.27+\shift+0.3-1.5, -5.9+0.15*3, 0.75*3) {\large$\mathbf{F}_1$};

	\pgfmathsetmacro{\shift}{1.6}
 
        \draw [myarrows, color = orange!80, dotted](12.4+\shift+0.3+2.1, -5.9+0.15*3, 0.75*3) -- (11.4+\shift+2.3, -5.9+0.15*3, 0.75*3);  
        
        \draw [myarrows, color = orange!80, dotted](12.4+\shift+0.3+2.1, -3.9+0.15*3, 0.75*3) -- (11.4+\shift+2.3, -3.9+0.15*3, 0.75*3);  
        
        \draw [myarrows, color = orange!80, dotted](12.4+\shift+0.3+2.1, -0.5+0.15*3, 0.75*3) -- (11.4+\shift+2.3, -0.5+0.15*3, 0.75*3);  
        
        \draw [myarrows, color = orange!80, dotted](12.4+\shift+0.3+2.1, 1.5+0.15*3, 0.75*3) -- (11.4+\shift+2.3, 1.5+0.15*3, 0.75*3);  
        
        \draw [mylines, color = orange!80, dotted](12.4+\shift+0.3+2.1, -5.9+0.15*3, 0.75*3) -- (12.4+\shift+0.3+2.1, 1.5+0.15*3, 0.75*3); 
        
        \draw [mylines, color = orange!80, dotted](12.4+\shift+0.3+2.1, -2.2+0.15*3, 0.75*3) -- (13.7+\shift+0.3+2.3, -2.2+0.15*3, 0.75*3); 
        

    \pgfmathsetmacro{\i}{16.8}
    \pgfmathsetmacro{\j}{-2.2}
    \draw[black,fill=gray!35, line width = 0.02mm] (\i+\shift,\j+0.15*3,0.1*3) -- ++(-\cubex,0,0) -- ++(0,-\cubey,0) -- ++(\cubex,0,0) -- cycle;
    \draw[black,fill=gray!35, line width = 0.02mm] (\i+\shift,\j+0.15*3,0.1*3) -- ++(0,0,-\cubez) -- ++(0,-\cubey,0) -- ++(0,0,\cubez) -- cycle;
    \draw[black,fill=gray!35, line width = 0.02mm] (\i+\shift,\j+0.15*3,0.1*3) -- ++(-\cubex,0,0) -- ++(0,0,-\cubez) -- ++(\cubex,0,0) -- cycle;

    \node at (16.5+\shift+0.1*3, -2.2+0.15*3+0.3, 0.75*3) {\large Task};
    \node at (16.5+\shift+0.1*3, -2.2+0.15*3-0.3, 0.75*3) {\large Labels};
    \node at (16.5+\shift+0.1*3, -3.5+0.15*3, 0.75*3) {\large${\bf{T}} = \{ T_1, \, ... \,, T_m  \}$};

    \pgfmathsetmacro{\dx}{6.9}
    \pgfmathsetmacro{\dy}{0}
    \draw [myarrows, color = matcha!150](-0.5+\dx, -7.35+\dy) -- (8+\dx, -7.35+\dy);  
    \node[anchor=north] at (3.7+\dx, -7.4+\dy) {\textcolor{matcha!150}{\large\textbf{Feature Channel}}};



		\pgfmathsetmacro{\x}{19.2}
		\pgfmathsetmacro{\y}{1.64}
		\pgfmathsetmacro{\dx}{1}
		\pgfmathsetmacro{\dya}{0.65} 
		\pgfmathsetmacro{\dyb}{0.52} 
		
		 \draw[thick, color = black] (\x-0.05, \y + 0.35) -- (29.3, \y + 0.35);
		 \node at (15.52, 2) {\hypertarget{alg}{\color{white}~}}; 
		\node at (\x+3.15, \y){\begin{tabular}{c}{\large {\bf{Alg. 1: }} Pseudocode for \texttt{BrainFM}}\end{tabular}};
		 \draw[thick, color = black] (\x-0.05, \y - 0.3) -- (29.3, \y - 0.3);
		 
		 \node at (\x + 2.84, \y - \dya - 0.02) {\begin{tabular}{c}{\large \color{olive!110}{\# $\mathcal{F}$: feature extraction network}} \end{tabular}};
		 \node at (\x + 2.42, \y - \dya - 0.9*\dyb- 0.02) {\small\begin{tabular}{c}{\large \color{olive!110}{\large \# $\mathcal{L}$: linear activation layer}} \end{tabular}};
		 \node at (\x + 4.69, \y - \dya - 1.8*\dyb- 0.02) {\small\begin{tabular}{c}{\large \color{olive!110}{\# $\{ \Theta_i\}_1^N$: generation params (contrast, resolution, ...)}} \end{tabular}};

		 \draw[thin, color = black] (\x + 0.68, \y - \dya - 3.5 * \dyb) -- (\x + 0.68, \y - \dya - 16. * \dyb)  -- (\x + 0.68 + 0.18, \y - \dya - 16. * \dyb) ;
   
		 \node at (\x + 0.2, \y - \dya - 3 * \dyb) {\begin{tabular}{c}{\small{\bf{1}}}\end{tabular}};
		 \node at (\x + 4.52, \y - \dya - 3 * \dyb) {\small\begin{tabular}{c}{\large \hypertarget{code: loader}{{\bf{for }}$(L,~T)$ in loader:~~\color{olive!110}{\# load 1 subject sample}}} \end{tabular}};
 
		 \node at (\x + 0.2, \y - \dya - 4 * \dyb) {\begin{tabular}{c}{\small{\bf{2}}}\end{tabular}};
		 \node at (\x + 4.58, \y - \dya - 4 * \dyb) {\small\begin{tabular}{c}{\large \color{olive!110}\# generate subject-level deformation field} \end{tabular}};
	   
		 \node at (\x + 0.2, \y - \dya - 5 * \dyb) {\begin{tabular}{c}{\hypertarget{code: 3}{\small{\bf{3}}}} \end{tabular}};
		 \node at (\x + 4.64, \y - \dya - 5 * \dyb) {\small\begin{tabular}{c}{\large $\phi = \textit{get}\_\textit{random}\_\textit{deformation()}$ via \cref{eq: deform}} \end{tabular}};
   
		 \node at (\x + 0.2, \y - \dya - 6 * \dyb) {\begin{tabular}{c}{\hypertarget{code: 4}{\small{\bf{4}}}} \end{tabular}};
		 \node at (\x + 4.03, \y - \dya - 6 * \dyb) {\small\begin{tabular}{c}{\large \color{olive!110}\# generate $N$ intra-subject samples} \end{tabular}};
		 
		 \node at (\x + 0.2, \y - \dya - 7 * \dyb) {\begin{tabular}{c}{\small{\bf{5}}} \end{tabular}};
		 \node at (\x + 2.58, \y - \dya - 7 * \dyb) {\small\begin{tabular}{c}{\large intra$\_$samples = []} \end{tabular}};

		 \draw[thin, color = black] (\x + 1.25, \y - \dya - 8.5 * \dyb) -- (\x + 1.25, \y - \dya - 10. * \dyb)  -- (\x + 1.25 + 0.18, \y - \dya - 10. * \dyb) ;
   
		 \node at (\x + 0.2, \y - \dya - 8 * \dyb) {\begin{tabular}{c}{\small{\bf{6}}} \end{tabular}};
		 \node at (\x + 3.99, \y - \dya - 8 * \dyb) {\small\begin{tabular}{c}{\large \textbf{for} $(i, ~\Theta_i)$ in enumerate$\big(\{\Theta_i\}_1^N\big)$:} \end{tabular}};
		 
		 \node at (\x + 0.2, \y - \dya - 9 * \dyb) {\begin{tabular}{c}{\hypertarget{code: 7}{\small{\bf{7}}}} \end{tabular}};
		 \node at (\x + 4.45, \y - \dya - 9 * \dyb) {\small\begin{tabular}{c}{\large $S_i = \textit{random}\_\textit{generate}(L\,|\,\phi, \,\Theta_i)$} \end{tabular}};
		 
		 \node at (\x + 0.2, \y - \dya - 10 * \dyb) {\begin{tabular}{c}{\hypertarget{code: 8}{\small{\bf{8}}}} \end{tabular}};
		 \node at (\x + 3.9, \y - \dya - 10 * \dyb) {\small\begin{tabular}{c}{\large intra$\_$samples.\textit{append}($S_i$)} \end{tabular}};
		 		 

		 \node at (\x + 0.2, \y - \dya - 11 * \dyb) {\begin{tabular}{c}{\hypertarget{code: 9}{\small{\bf{9}}}} \end{tabular}};
		\node at (\x + 4.42, \y - \dya - 11 * \dyb) {\small\begin{tabular}{c}{\large \color{olive!110}\# formulate all samples to a mini-batch} \end{tabular}};

		 \node at (\x + 0.14, \y - \dya - 12 * \dyb) {\begin{tabular}{c}{\hypertarget{code: 10}{\small{\bf{10}}}} \end{tabular}};
		 \node at (\x + 3.3, \y - \dya - 12 * \dyb) {\small\begin{tabular}{c}{\large $\mathbf{x} = concat(\text{intra}\_\text{samples})$} \end{tabular}};
		 
		 \node at (\x + 0.144, \y - \dya - 13 * \dyb) {\begin{tabular}{c}{\small{\bf{11}}} \end{tabular}};
		 \node at (\x + 3.7, \y - \dya - 13 * \dyb) {\small\begin{tabular}{c}{\large $\mathbf{F} = \mathcal{F}(\mathbf{x})$~~{\color{olive!110}\# feature extraction}} \end{tabular}}; 
		 
		 \node at (\x + 0.14, \y - \dya - 14 * \dyb) {\begin{tabular}{c}{\small{\bf{12}}} \end{tabular}};
		 \node at (\x + 4.45, \y - \dya - 14 * \dyb) {\small\begin{tabular}{c}{\large {\color{olive!110}\# subject-specific multi-task supervision}} \end{tabular}};

		 \node at (\x + 0.14, \y - \dya - 15 * \dyb) {\begin{tabular}{c}{\small{\bf{13}}} \end{tabular}};
		 \node at (\x + 3.32, \y - \dya - 15 * \dyb) {\small\begin{tabular}{c}{\large loss = $\sum_{i=1}^m L_{i}(\phi_i(\mathbf{F}),~T_i)$} \end{tabular}};
		  
		 \node at (\x + 0.14, \y - \dya - 16 * \dyb) {\begin{tabular}{c}{\small{\bf{14}}} \end{tabular}};
		 \node at (\x + 2.43, \y - \dya - 16 * \dyb) {\small\begin{tabular}{c}{\large loss$.backward()$} \end{tabular}};
		 
		 \draw[thick, color = black] (\x-0.05, \y - \dya - 16.7 * \dyb) -- (29.3, \y - \dya - 16.7 * \dyb);

	\end{tikzpicture}
	}   
\caption{\texttt{BrainFM}'s modality-agnostic multi-task learning framework.}  
	 \label{fw: train}
\end{figure*}

%% file: sec/method/trainer.tex
\subsection{Modality-agnostic Multi-task Learning}
\label{sec: framework}

As mentioned above, we would like the resulting features from \texttt{BrainFM} to be robust to intra-subject variations \textit{and} expressive to potential downstream tasks. In this section, we introduce \texttt{BrainFM}'s learning framework to achieve these two desired properties.


\input{sec/exp/fig/fig_comp}

\subsubsection{Intra-subject Data Generation}
\label{sec: sample}
In order to learn a feature representation that is distinctive to each subject and robust to varying MR contrasts, we observe that enriching \textit{intra}-subject samples leads to better performance~(\cref{tab: ablat}). Specifically, instead of including multiple subjects for a mini-batch during each training iteration as in usual practice, \texttt{BrainFM} focuses on maximizing the intra-subject variance to improve the subject-robustness of resulting features. As described in Alg.~\hyperlink{alg}{1} (line \hyperlink{code: 3}{4}-\hyperlink{code: 9}{10}), for each training iteration, after randomly selecting a subject and generating its deformation, \texttt{BrainFM} generates a mini-batch of intra-subject samples ($\{ S_1, \, \dots, \, S_N \}$) with randomly synthesized contrasts, resolutions and corruptions~(\cref{sec: generator}). \texttt{BrainFM} collects losses from all intra-subject samples and conducts back-propagation \textit{at once},  to encourage the subject-specific robustness of its learned features.

We set the intra-subject samples within a mini-batch to have random contrasts and ``mild-to-severe'', \textit{increasing} level of corruptions (\cref{fw: train} (left)), to maximize the intra-subject variance while ensuring the stability of the training process against extreme corruption levels. As shown in Fig.~\ref{fig: feat_comp}, \texttt{BrainFM} obtains contrast/resolution-robustness that \textit{cannot} be achieved by models trained from real images (due to the \textit{limited variability} in their appearance), regardless of the backbone choices.


\subsubsection{Multi-task Supervision}
\label{sec: tasks}

\paragraph{Image Synthesis}
\label{task: synthesis}~For each training subject, if the ground truth modality is available, we train models to reconstruct their T1w MRI, T2w MRI, FLAIR, and CT images with the \texttt{L1} loss and gradient \texttt{L1} loss.

\paragraph{Atlas Registration}
\label{task: reg}~To achieve a unified framework for registration, we adopt registration by regression (R2R), an atlas registration framework which predicts the atlas coordinates for every voxel of the input scan (i.e., every voxel is a keypoint)~\cite{gopinath2024rbr,hu2025hierarchical}. We follow R2R with the \texttt{L1} loss and gradient \texttt{L1} loss.

\paragraph{Anatomy Segmentation}
\label{task: seg}~For all training subjects, we use SynthSeg~\cite{Billot2021SynthSegSO} to obtain the segmentation labels with 30 brain anatomical regions~\cite{Fischl2002WholeBS}, as the gold standard segmentation target. We train models to predict the brain segmentation labels, with the soft dice loss and cross-entropy loss~\cite{Billot2021SynthSegSO}. 

\paragraph{Distance Maps Prediction}
\label{task: dist}~We predict distance maps to represent extracted surfaces, where each voxel encodes the shortest distance to the segmented boundary~\cite{gopinath2025recon}. All the distance values are trained with \texttt{L1} loss.

\paragraph{Bias Field Estimation}
\label{task: bf}~The bias field is a smooth,  low-frequency multiplicative signal that corrupts MRI images, which affects image analysis tasks such as segmentation or texture analysis~\cite{Juntu2005BiasFC}. Bias field estimation is often needed as a pre-processing step to correct corrupted MRI images~\cite{Chua2009EvaluationOP}. We apply randomly simulated bias fields to the input samples~(\cref{sec: corrupt}), and train models with the \texttt{L2} loss. Note that we pre-generate and apply the bias fields on the testing data for reproducibility, and evaluate the bias field estimation performance during inference.

%% file: sec/exp/fig/fig_comp.tex
\begin{figure}[t]

\centering 

\resizebox{\linewidth}{!}{
	\begin{tikzpicture}
        
		\tikzstyle{myarrows}=[line width=0.6mm,draw=blue!50,-triangle 45,postaction={draw, line width=0.05mm, shorten >=0.02mm, -}]
		\tikzstyle{mylines}=[line width=0.8mm]
  

	\pgfmathsetmacro{\shift}{-1.}
 
        \node at (\shift+1.05, 3+0.35,0.1*3) {\hypertarget{fig: feat_dist-a}{~}};
 
	\foreach \i in {-1, 1, 3}
	{
        \draw [<->, color = matcha!150, line width = 0.5mm](\shift+0.6,\i-0.15,0.1*3) -- (\shift+1.5,\i-0.15,0.1*3);  
	}

        
        \node at (0.+\shift+0.5-1.9, 3.2+0.15*3, 0.75*3) {T1w};
        \node at (0.+\shift+0.5-1.9, 2.8+0.15*3, 0.75*3) {(HF)};
        \node at (0.+\shift+0.5-1.9, 1.2+0.15*3, 0.75*3) {T1w};
        \node at (0.+\shift+0.5-1.9, 0.8+0.15*3, 0.75*3) {(\textbf{\textit{LF}})}; 
        \node at (0.+\shift+0.5-1.9, -0.8+0.15*3, 0.75*3) {\textbf{\textit{FLAIR}}};
        \node at (0.+\shift+0.5-1.9, -1.2+0.15*3, 0.75*3) {(HF)};

	\node at (\shift+0.1*3, 3+0.15*3, 0.75*3) {\includegraphics[width=0.12\textwidth]{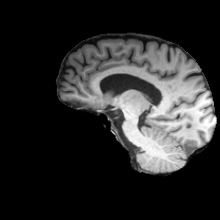}};
	\node at (\shift+0.1*3, 1+0.15*3, 0.75*3) {\includegraphics[width=0.12\textwidth]{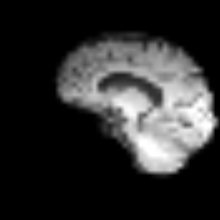}};
	\node at (\shift+0.1*3, -1+0.15*3, 0.75*3) {\includegraphics[width=0.12\textwidth]{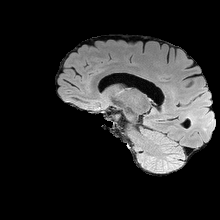}};

 
 
	\pgfmathsetmacro{\shift}{-3.2}

	\node at (5.3+\shift+0.1*3, 3+0.15*3, 0.75*3) {\includegraphics[width=0.12\textwidth]{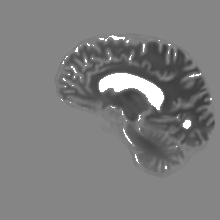}};
	\node at (5.3+\shift+0.1*3, 1+0.15*3, 0.75*3) {\includegraphics[width=0.12\textwidth]{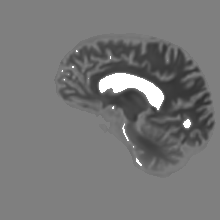}};
	\node at (5.3+\shift+0.1*3, -1+0.15*3, 0.75*3) {\includegraphics[width=0.12\textwidth]{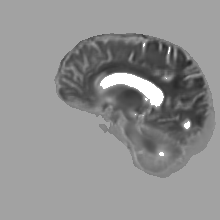}};
        \node at (5.3+\shift+0.1*3-0.5, 3-0.45+0.15*3, 0.75*3) {\textcolor{green}{\LARGE\cmark}};
        \node at (5.3+\shift+0.1*3-0.5, 1-0.45+0.15*3, 0.75*3) {\textcolor{green}{\LARGE\cmark}};
        \node at (5.3+\shift+0.1*3-0.5, -1-0.45+0.15*3, 0.75*3) {\textcolor{green}{\LARGE\cmark}};
        
	\node at (7.3+\shift+0.1*3, 3+0.15*3, 0.75*3) {\includegraphics[width=0.12\textwidth]{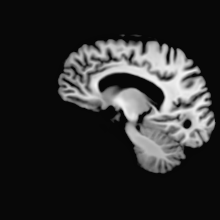}};
	\node at (7.3+\shift+0.1*3, 1+0.15*3, 0.75*3) {\includegraphics[width=0.12\textwidth]{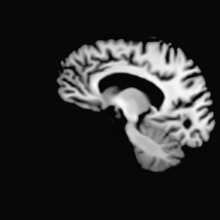}};
	\node at (7.3+\shift+0.1*3, -1+0.15*3, 0.75*3) {\includegraphics[width=0.12\textwidth]{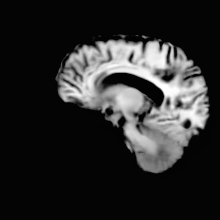}};
        \node at (7.3+\shift+0.1*3-0.5, 3-0.45+0.15*3, 0.75*3) {\textcolor{green}{\LARGE\cmark}};
        \node at (7.3+\shift+0.1*3-0.5, 1-0.45+0.15*3, 0.75*3) {\textcolor{green}{\LARGE\cmark}};
        \node at (7.3+\shift+0.1*3-0.5, -1-0.45+0.15*3, 0.75*3) {\textcolor{green}{\LARGE\cmark}};
 
	\node at (9.8+\shift+0.1*3, 3+0.15*3, 0.75*3) {\includegraphics[width=0.12\textwidth]{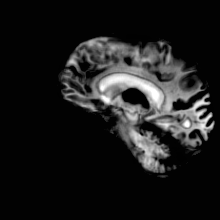}};
	\node at (9.8+\shift+0.1*3, 1+0.15*3, 0.75*3) {\includegraphics[width=0.12\textwidth]{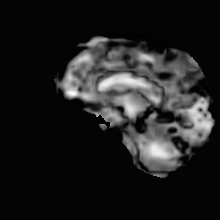}};
	\node at (9.8+\shift+0.1*3, -1+0.15*3, 0.75*3) {\includegraphics[width=0.12\textwidth]{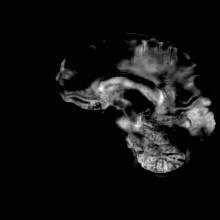}};
        \node at (9.8+\shift+0.1*3-0.5, 3-0.45+0.15*3, 0.75*3) {\textcolor{green}{\LARGE\cmark}};
        \node at (9.8+\shift+0.1*3-0.5, 1-0.45+0.15*3, 0.75*3) {\textcolor{red}{\LARGE\xmark}};
        \node at (9.8+\shift+0.1*3-0.5, -1-0.45+0.15*3, 0.75*3) {\textcolor{red}{\LARGE\xmark}};
 
	\node at (11.8+\shift+0.1*3, 3+0.15*3, 0.75*3) {\includegraphics[width=0.12\textwidth]{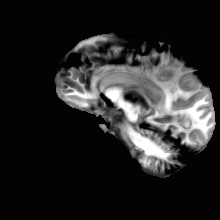}};
	\node at (11.8+\shift+0.1*3, 1+0.15*3, 0.75*3) {\includegraphics[width=0.12\textwidth]{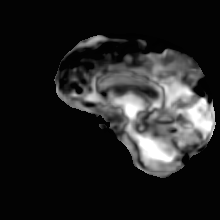}};
	\node at (11.8+\shift+0.1*3, -1+0.15*3, 0.75*3) {\includegraphics[width=0.12\textwidth]{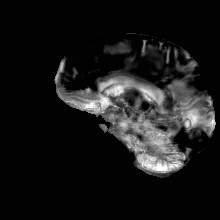}};
        \node at (11.8+\shift+0.1*3-0.5, 3-0.45+0.15*3, 0.75*3) {\textcolor{green}{\LARGE\cmark}};
        \node at (11.8+\shift+0.1*3-0.5, 1-0.45+0.15*3, 0.75*3) {\textcolor{red}{\LARGE\xmark}};
        \node at (11.8+\shift+0.1*3-0.5, -1-0.45+0.15*3, 0.75*3) {\textcolor{red}{\LARGE\xmark}};

    \pgfmathsetmacro{\dx}{-7.65}
    \pgfmathsetmacro{\dy}{5.28}
        \node[anchor=north] at (6.1+\dx, -7.65+\dy) {\small {\textbf{Input}}}; 
        
    \pgfmathsetmacro{\dy}{4.8}
        \node[anchor=north] at (6.1+\dx, -7.6+\dy) {\small (\textbf{HF}: high-field; \textbf{LF}: low-field)};

    \pgfmathsetmacro{\dx}{-3.5}
        \draw [decorate,decoration={brace,amplitude=5pt,mirror,raise=6ex},line width=2.pt,color = matcha!150] (3.9+\dx, -1.6) -- (8.2+\dx, -1.6);
        \node[anchor=north] at (6.1+\dx, -7.6+\dy) {\textcolor{matcha!150}{\small \textbf{\texttt{BrainFM} features}}};

    \pgfmathsetmacro{\dx}{1.}
        \draw [decorate,decoration={brace,amplitude=5pt,mirror,raise=6ex},line width=2.pt,color = gray!80] (3.9+\dx, -1.6) -- (8.2+\dx, -1.6);
        \node[anchor=north] at (6.1+\dx, -7.6+\dy) {\small \textcolor{gray!80}{\textbf{\texttt{SCRATCH} features}}}; 


	\end{tikzpicture}
	}  
\caption{\texttt{SCRATCH}, well trained on high-field T1w scans with the same model architecture as \texttt{BrainFM}, produces highly descriptive features for high-field T1w images ($1^{\text{st}}$ row), but \textit{does not preserve} the same high quality useful for downstream tasks when handling low-field ($2^{\text{nd}}$ row) or other contrasts ($3^{\text{rd}}$ row).} 
	 \label{fig: feat_comp}
  
\end{figure}
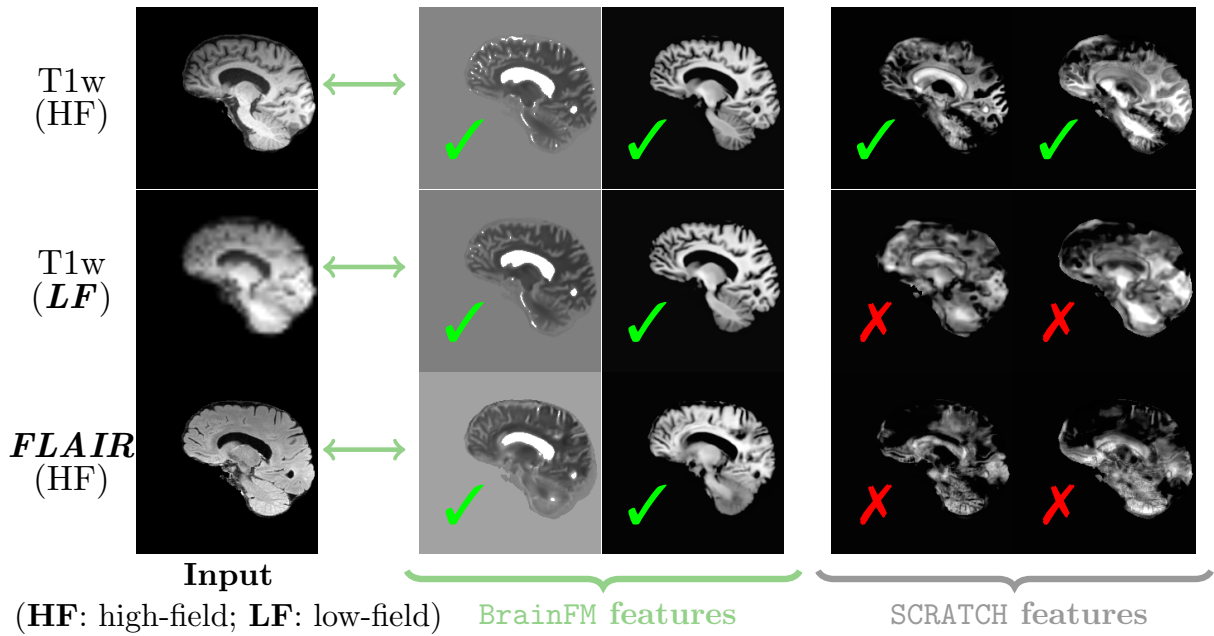

%% file: sec/method/fig/fw_multitask.tex
\begin{figure*}[t] 
\centering
\resizebox{0.85\textwidth}{!}{
\begin{tikzpicture}

\node at (0, 0) {\includegraphics[width=\linewidth]{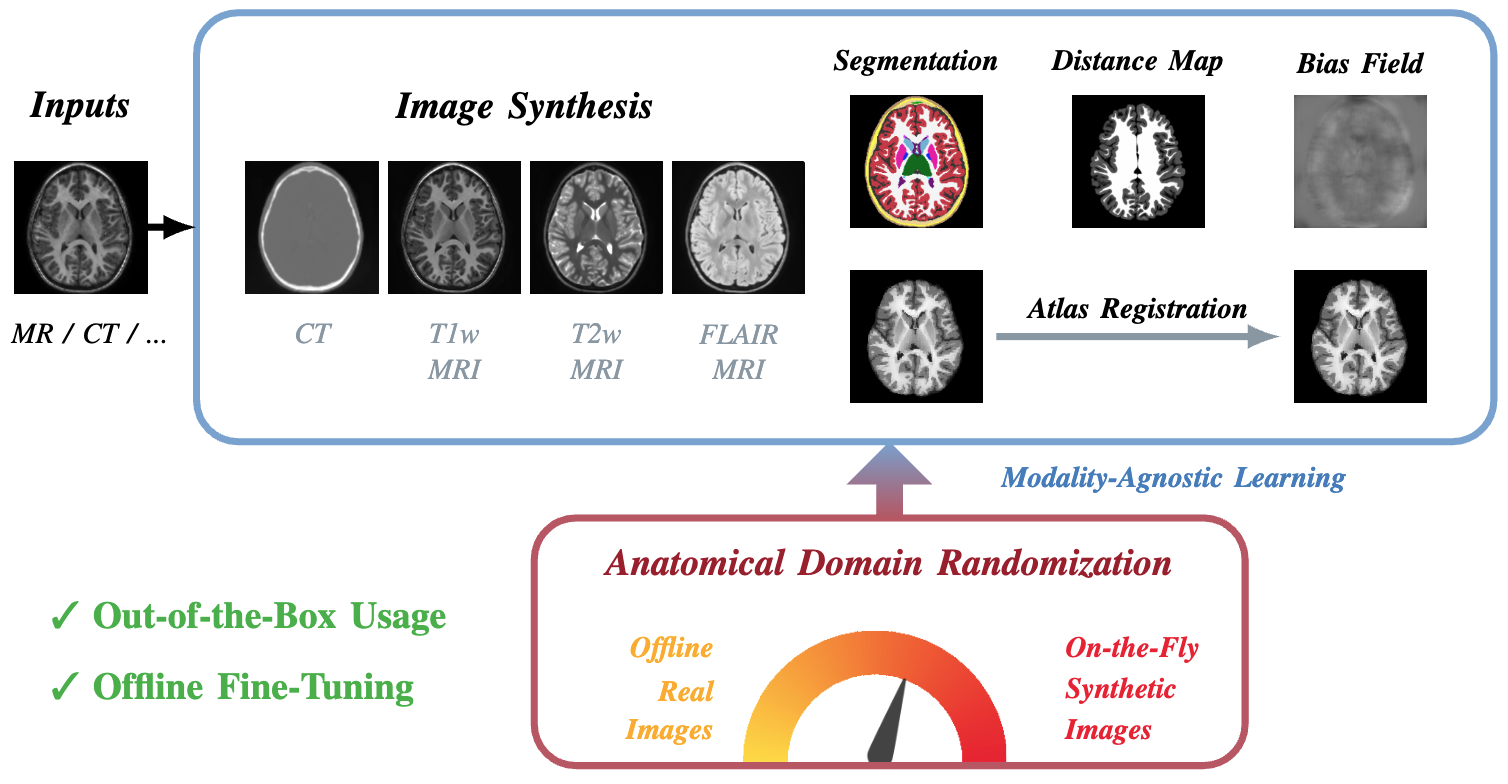}}; 
\end{tikzpicture}
}
\caption{\texttt{BrainFM}'s modality-agnostic multi-task software.}  
	 \label{fw: multitask}
\end{figure*}

%% file: sec/exp/main.tex

\section{Experiments}
\label{sec: exp}


\input{sec/exp/setup}

\input{sec/exp/upstream} 

\input{sec/exp/additional}


%% file: sec/exp/setup.tex
\subsection{Datasets, Metrics, Models, Implementation Details}
\label{sec: setup}


\subsubsection{Datasets}
\label{sec: datasets}
We experience \texttt{BrainFM} over eleven datasets including modalities of MR and CT, the MR images further contain T1-weighted, T2-weighted, and FLAIR (fluid-attenuated inversion recovery) images. 



\begin{itemize}

\item \texttt{ABIDE}~\cite{di2014abide,di2017abide}: T1-weighted (819 cases) MRI scans from the Autism Brain Imaging Data Exchange, an initiative which has aggregated functional and structural brain imaging data collected from laboratories around the world for the understanding of the neural bases of autism.

\item \texttt{ADHD200}~\cite{Brown2012ADHD200GC}: T1-weighted (820 cases) MRI scans from ADHD200 Sample, a grassroots initiative dedicated to the understanding of the neural basis of Attention Deficit Hyperactivity Disorder (ADHD). We also use \texttt{ADHD200} for the downstream brain age estimation task.

\item \texttt{ADNI3}~\cite{Weiner2017TheAD}: T1-weighted (316 cases) and FLAIR (315 cases) MRI scans from ADNI3, which continues the previously funded ADNI1, ADNI-GO, and ADNI2 studies to determine the relationships between the clinical, cognitive, imaging, genetic and biochemical biomarker characteristics of the entire spectrum of sporadic late-onset AD.

\item \texttt{AIBL}~\cite{Fowler2021FifteenYO}: T1-weighted (646 cases), T2-weighted (302 cases), and FLAIR (273 cases) MRI scans from the Australian Imaging, Biomarkers and Lifestyle (AIBL) Study, for cognitive impairment (MCI) and Alzheimer’s disease dementia.

\item \texttt{Buckner40}: a subset of a larger open-access structural data set created by the Buckner lab\footnote{\url{https://bucknerlab.fas.harvard.edu/data-tools}}. It consists of 38 T1-weighted MRI scans. 

\item \texttt{COBRE}~\cite{calhoun2012exploring}: T1-weighted (187 cases) MRI scans from the Center for Biomedical Research Excellence, including raw anatomical and functional MR data from 72 patients with Schizophrenia and 75 healthy controls.

\item \texttt{ISBI2015}~\cite{Carass2017ISBI2015}: T1-weighted (21 cases) MRI scans from the Longitudinal MS Lesion Segmentation Challenge for ISBI 2015.

\item \texttt{HCP}~\cite{Essen2012TheHC}: T1-weighted (897 cases) and T2-weighted (897 cases) healthy MRI scans of young subjects from the Human Connectome Project, acquired at 0.7 mm resolution.

\item \texttt{Chinese-HCP}~\cite{Vogt2012ChineseHCP}: T1-weighted (212 cases) healthy MRI scans from the Chinese Human Connectome Project, a large-scale neuroimaging project parallel to the HCP, but with a focus on the East Asian population.

\item \texttt{MCIC}~\cite{gollub2013mcic}: T1-weighted (161 cases) MRI scans from the schizophrenic and matched control data collection, consisting of images from schizophrenia patients and gender and age-matched controls.

\item \texttt{OASIS3}~\cite{LaMontagne2018OASIS3LN}: T1-weighted MRI (1238 cases), T2-weighted MRI (695 cases), and CT (838 cases) scans from OASIS3, which is a longitudinal neuroimaging, clinical, and cognitive dataset for normal aging and AD. We used T1-weighted MRI, T1-weighted MRI, and CT pairs with the earliest date for each subject.

\end{itemize}

\input{sec/exp/fig/tab_data}


Considering the potential domain gaps across datasets,  we experiment with three train/test setups, to evaluate their capability better. As detailed in Tab.~\ref{tab: data}. (1) \texttt{Setup-\textit{I}}, the training data includes samples from all datasets, with 10\% images of each dataset randomly selected for testing; (2) \texttt{Setup-\textit{II}}, the training data is from all datasets except \texttt{HCP} and \texttt{ADNI3}, that are left for testing performance on T2w and FLAIR MR. Since \texttt{OASIS3} is the only dataset that includes CT images, we use the same train/test split as in \texttt{Setup-\textit{I}}.; (3) \texttt{Setup-\textit{III}}, in complement to \texttt{Setup-\textit{II}}, \texttt{Setup-\textit{III}} includes training data from all datasets except \texttt{AIBL}, which is left for performance testing on FLAIR. For CT, we also use the same train/test split as in \texttt{Setup-\textit{I}}.

\subsubsection{Metrics}
\label{sec: metrics}
We evaluate individual tasks from different aspects. For image synthesis and super-resolution, we use \texttt{L1}, \texttt{PSNR} (peak signal-to-noise ratio) and \texttt{SSIM}. For segmentation, we use Dice scores. For atlas registration and distance map estimation, we use \texttt{L1}. For bias field estimation, we use the normalized \texttt{L2} distance (\texttt{norm-L2}) to avoid possible arbitrary scalings from nonuniformity correction~\cite{Chua2009EvaluationOP}.


\subsubsection{Models for Comparison}\label{sec: models}

We compare performance of the above \texttt{BrainFM} variants with: \textit{(i-ii)}~\texttt{SAMSEG}~\cite{Puonti2016FastAS,Cerri2020ACM} and \texttt{FastSurfer}~\cite{Henschel2022FastSurferVINN} (only works on T1w), state-of-the-art classical and machine-learning-based brain segmentation models, respectively; \textit{(iii)}~\texttt{SynthSR}~\cite{Iglesias2023SynthSRAP}, state-of-the-art, contrast-agnostic T1w synthesis model. We also provide fine-tuned \texttt{SynthSR} (\texttt{SynthSR}-FT) results for further comparison.


\subsubsection{Data Preprocessing}\label{sec: preprocess} For all datasets, we skull-strip all the images using SynthStrip~\cite{Hoopes2022SynthStripSF}, and resample them to 1 $mm$ isotropic resolution. For all the images, except T1-weighted MRI, in each dataset, we use NiftyReg~\cite{Modat2010FastFD} rigid registration to register all images to their same-subject T1-weighted MRI counterparts. The gold-standard brain segmentation maps are obtained by performing SynthSeg~\cite{Billot2021SynthSegSO} on the T1-weighted MR images of all the subjects. For consistency across datasets, we further masked out the de-faced regions for the datesets.

\texttt{BrainFM}'s synthetic generator uses (1) brain segmentation labels, for random-contrast input images generation~(\cref{sec: generator}), and (2) randomly chosen real images for ``real-synth'' mix-up~(Sec.~\ref{sec: mixup}).


\subsubsection{Training}\label{sec: implement} As a general feature representation model, \texttt{BrainFM} can use any backbone to extract brain features. For fairer comparison, we adopt the same five-level 3D UNet~\cite{Ronneberger2015UNetCN} (with 64 feature channels in the last layer). A linear regression layer is then added for each task. The training is conducted in an end-to-end fashion. We co-train \texttt{BrainFM} on a combination of all the real datasets listed in Sec.~\ref{sec: datasets} and the synthetic images our generator (\cref{sec: generator}), for 2,000 epochs, with a patch size of $128^3$ and a mini-batch size (i.e., number of intra-subject augmented samples) of 4. We use the synchronized AdamW optimization, with a base learning rate of $10^{-4}$ and a linear warm-up in the first 5 epochs, followed by a multi-step learning schedule (learning rate drops at 1,000 and 1,600 epochs) with a multiplier of 0.1.

%% file: sec/exp/fig/tab_data.tex
\begin{table}[t]
    \caption{Details on the available modalities/contrasts of each dataset, and the training/testing splits of \texttt{BrainFM} variants.} 
    \label{tab: data}
\resizebox{0.7\linewidth}{!}{
\centering 
    \begin{tabular}{clcccc}
       \toprule \\[-3ex] 
      \multicolumn{1}{c}{\multirow{2}{*}{\bf Setup}} &  \multicolumn{1}{c}{\multirow{2}{*}{\bf Split}} & \multicolumn{3}{c}{\textbf{MR}} & \multicolumn{1}{c}{\multirow{2}{*}{\bf CT}} \\ [-0.4ex]
        \cmidrule(lr){3-5}
           & & T1w & T2w & FLAIR \\ [-0.2ex]
     \midrule\\[-2.8ex]
      
        &  \multicolumn{1}{c}{\multirow{2}{*}{\it Availablity}} & {\multirow{2}{*}{All}} & \texttt{HCP} & \texttt{ADNI3} & {\multirow{2}{*}{\texttt{OASIS3}}}  \\ 
         & & & \texttt{OASIS3} & \texttt{AIBL} &   \\ 
         \midrule\\[-3.4ex]
     \midrule\\[-2.8ex]
         
        {\multirow{4}{*}{\texttt{\textit{I}}}} & {\multirow{2}{*}{\it train}} & {\multirow{2}{*}{90\% All}} & 90\% \texttt{HCP} & 90\% \texttt{ADNI3} & {\multirow{2}{*}{90\% \texttt{OASIS3}}} \\ 
          &   &  & 90\% \texttt{OASIS3} & 90\% \texttt{AIBL} &   \\ 
           \cmidrule(lr){2-6}
         & {\multirow{2}{*}{\it test}} & {\multirow{2}{*}{10\% All}} & 10\% \texttt{HCP} & 10\% \texttt{ADNI3} & {\multirow{2}{*}{10\% \texttt{OASIS3}}} \\ 
          &   &  & 10\% \texttt{OASIS3} & 10\% \texttt{AIBL} &   \\ 
         \hline\\[-2.3ex]

        {\multirow{3}{*}{\texttt{\textit{II}}}} & {\multirow{2}{*}{\it {train}}} & \multicolumn{1}{c}{\multirow{2}{*}{\thead{All except\\ \texttt{HCP}, \texttt{ADNI3}}}}
        &  {\multirow{2}{*}{\texttt{OASIS3}}} & {\multirow{2}{*}{\texttt{AIBL}}} & {\multirow{2}{*}{90\% \texttt{OASIS3}}} \\ [-0.7ex]
          &   &  &   &   &   \\ 
           \cmidrule(lr){2-6}
         & {\it test} & \texttt{HCP}, \texttt{ADNI3} & \texttt{HCP} & \texttt{ADNI3} & 10\% \texttt{OASIS3} \\ 
         \hline\\[-1.7ex]

        {\multirow{3}{*}{\texttt{\textit{III}}}} & {\multirow{2}{*}{\it train}} & \multicolumn{1}{c}{\multirow{2}{*}{\thead{All except\\ \texttt{AIBL}}}}
        &  {\multirow{2}{*}{\texttt{HCP}}} & {\multirow{2}{*}{\texttt{ADNI3}}} & {\multirow{2}{*}{90\% \texttt{OASIS3}}} \\ [-0.7ex]
          &   &  &   &   &   \\ 
           \cmidrule(lr){2-6}
         & {\it test} & \texttt{AIBL} & \texttt{OASIS3} & \texttt{AIBL} & 10\% \texttt{OASIS3} \\ 
\bottomrule  \\ [-3.6ex]  
    \end{tabular} 
}
\end{table}

%% file: sec/exp/upstream.tex
\subsection{Tasks Evaluation}
\label{exp: upstream}

\input{sec/exp/fig/tab_recon}
\input{sec/exp/fig/tab_other_tasks} 

In \texttt{Brain-ID}~\cite{Liu2024BrainID}, we have validated the robustness and expressiveness properties of our anatomical domain randomization design. In this work, we particularly investigate \texttt{BrainFM}'s multi-task performance (Sec.~\ref{sec: tasks}) against \textit{(i)}~\texttt{Brain-ID} on the five tasks described in Sec .~\ref{sec: tasks}, to demonstrate the advantages of ``real-synth'' mix-up and multi-task training in \texttt{BrainFM}; \textit{(ii)}~\texttt{SCRATCH}: a baseline with the same data generation/architecture as \texttt{Brain-ID} and \texttt{BrainFM}, yet trained from \textit{scratch}. \texttt{SCRATCH} is used to validate \texttt{Brain-ID}'s effectiveness;


Table~\ref{tab: recon} shows the results for image synthesis experiments, where \texttt{BrainFM} consistently demonstrates superior reconstruction performance over both \texttt{SCRATCH} and \texttt{Brain-ID} across all imaging modalities and training setups. Compared to \texttt{SCRATCH}, \texttt{BrainFM} markedly reduces \texttt{L1} errors while substantially improving \texttt{PSNR} and \texttt{SSIM}, confirming the effectiveness of leveraging anatomical priors rather than training from scratch. Moreover, \texttt{BrainFM} achieves consistent gains over \texttt{Brain-ID}, underscoring the benefit of combining real and synthetic data during training as well as exploiting the multi-task learning setup. These improvements are evident across T1w, T2w, FLAIR MR, and CT images, suggesting that \texttt{BrainFM} is not only more accurate in capturing structural information, but also more robust to variations in modality. The fact that \texttt{BrainFM} performs reliably across three distinct experimental setups further highlights its generalizability and adaptability to heterogeneous acquisition conditions, a crucial property for practical medical imaging pipelines.

Beyond synthesis, \texttt{BrainFM} also provides consistent improvements in a range of downstream tasks, as shown in Table~\ref{tab: other_tasks}. For atlas registration and distance prediction, \texttt{BrainFM} achieves the lowest \texttt{L1} errors, which translate into more precise spatial correspondences and more accurate geometric reasoning within brain anatomy. In segmentation, \texttt{BrainFM} yields the highest \texttt{Dice} scores across modalities, indicating stronger anatomical boundary delineation and more faithful recovery of fine structures compared to both baselines. For bias field estimation, \texttt{BrainFM} either matches or surpasses \texttt{Brain-ID} with lower \texttt{norm-L2} errors, reflecting more reliable correction of scanner-induced intensity inhomogeneities. Collectively, these results show that \texttt{BrainFM}’s advantages extend beyond pure generative fidelity to tasks directly supporting quantitative neuroimaging analyses. The consistent performance gains suggest that the combination of real-synth mix-up and multi-task training fosters representations that are both data-efficient and broadly transferable, ultimately enabling a more versatile and effective framework for brain image modeling.




%% file: sec/exp/fig/tab_recon.tex
\begin{table}[t]
    \caption{Comparisons of \texttt{BrainFM} with \texttt{SCRATCH} and \texttt{Brain-ID}, on image synthesis. Corresponding train/test setups are listed in Tab.~\ref{tab: data}. Each entry reports the metric values for \texttt{L1} ($\downarrow$) / \texttt{PSNR} ($\uparrow$) / \texttt{SSIM} ($\uparrow$).} 
    \label{tab: recon}
\resizebox{0.7\linewidth}{!}{
\centering 
    \begin{tabular}{ccccccc}
       \toprule \\[-3ex] 
      \multicolumn{1}{c}{\multirow{2}{*}{\bf Setup}} &  \multicolumn{1}{c}{\multirow{2}{*}{\bf Model}} & \multicolumn{1}{c}{\multirow{2}{*}{\bf \backslashbox{Input}{Output}}} & \multicolumn{3}{c}{\textbf{MR}} & \multicolumn{1}{c}{\multirow{2}{*}{\bf CT}} \\ [-0.4ex]
        \cmidrule(lr){4-6}
           & & & T1w & T2w & FLAIR \\ [-0.6ex]
     \midrule\\[-2.8ex]
       
        {\multirow{12}{*}{\texttt{\textit{I}}}} & \texttt{SCRATCH} & {\multirow{3}{*}{T1w MR}} & 0.025 & 0.033 & 0.048 & 70.12 \\
         & \texttt{Brain-ID} &  & 0.017 & 0.025 & 0.039 & 67.15 \\ 
          & \texttt{BrainFM}  &  &  \textbf{0.015} & \textbf{0.021} & \textbf{0.036} & \textbf{60.89}   \\ 
           \cmidrule(lr){2-7} 
         &  \texttt{SCRATCH} & {\multirow{3}{*}{T2w MR}} & 0.038 & 0.021 & 0.036 & 63.10 \\
         &  \texttt{Brain-ID} &  & 0.031 & 0.016 & 0.030 & 60.35  \\ 
          & \texttt{BrainFM}  &  &  \textbf{0.025} & \textbf{0.014} & \textbf{0.026} & \textbf{56.26}   \\ 
           \cmidrule(lr){2-7} 
         &  \texttt{SCRATCH} & {\multirow{3}{*}{FLAIR MR}} & 0.034 & 0.036 & 0.032 & - \\
         &  \texttt{Brain-ID} &  & 0.028 & 0.029 & 0.026 & - \\ 
          & \texttt{BrainFM}  &  &  \textbf{0.025} & \textbf{0.026} & \textbf{0.021} & -   \\ 
           \cmidrule(lr){2-7} 
         &  \texttt{SCRATCH} & {\multirow{3}{*}{CT}} & 0.075 & 0.052 & - & 40.25 \\
         &  \texttt{Brain-ID} &  & 0.061 & 0.045 & - & 37.68 \\ 
          & \texttt{BrainFM}  &  &  \textbf{0.050} & \textbf{0.041} & - & \textbf{34.51}   \\ 
         \hline\\[-2.3ex]

        {\multirow{12}{*}{\texttt{\textit{II}}}} & \texttt{SCRATCH} & {\multirow{3}{*}{T1w MR}} & 0.026 & 0.035 & 0.049 & 71.02 \\
         & \texttt{Brain-ID} &  & 0.019 & 0.028 & 0.040 & 67.31 \\ 
          & \texttt{BrainFM}  &  &  \textbf{0.016} & \textbf{0.023} & \textbf{0.036} & \textbf{60.17}   \\ 
           \cmidrule(lr){2-7} 
         &  \texttt{SCRATCH} & {\multirow{3}{*}{T2w MR}} & 0.039 & 0.023 & 0.037 & 64.25 \\
         &  \texttt{Brain-ID} &  & 0.033 & 0.018 & 0.031 & 61.02  \\ 
          & \texttt{BrainFM}  &  &  \textbf{0.027} & \textbf{0.015} & \textbf{0.028} & \textbf{56.92}   \\ 
           \cmidrule(lr){2-7} 
         &  \texttt{SCRATCH} & {\multirow{3}{*}{FLAIR MR}} & 0.035 & 0.038 & 0.033 & - \\
         &  \texttt{Brain-ID} &  & 0.030 & 0.031 & 0.028 & - \\ 
          & \texttt{BrainFM}  &  &  \textbf{0.025} & \textbf{0.026} & \textbf{0.022} & -   \\ 
           \cmidrule(lr){2-7} 
         &  \texttt{SCRATCH} & {\multirow{3}{*}{CT}} & 0.076 & 0.054 & - & 41.11 \\
         &  \texttt{Brain-ID} &  & 0.062 & 0.047 & - & 38.72 \\ 
          & \texttt{BrainFM}  &  &  \textbf{0.052} & \textbf{0.043} & - & \textbf{35.10}   \\ 
         \hline\\[-2.3ex]

        {\multirow{12}{*}{\texttt{\textit{III}}}} & \texttt{SCRATCH} & {\multirow{3}{*}{T1w MR}} & 0.027 & 0.034 & 0.051 & 71.38 \\
         & \texttt{Brain-ID} &  & 0.019 & 0.026 & 0.042 & 67.59 \\ 
          & \texttt{BrainFM}  &  &  \textbf{0.015} & \textbf{0.022} & \textbf{0.038} & \textbf{60.40}   \\ 
           \cmidrule(lr){2-7} 
         &  \texttt{SCRATCH} & {\multirow{3}{*}{T2w MR}} & 0.040 & 0.022 & 0.038 & 64.80 \\
         &  \texttt{Brain-ID} &  & 0.033 & 0.017 & 0.033 & 60.85  \\ 
          & \texttt{BrainFM}  &  &  \textbf{0.028} & \textbf{0.014} & \textbf{0.029} & \textbf{57.00}   \\ 
           \cmidrule(lr){2-7} 
         &  \texttt{SCRATCH} & {\multirow{3}{*}{FLAIR MR}} & 0.036 & 0.038 & 0.034 & - \\
         &  \texttt{Brain-ID} &  & 0.030 & 0.031 & 0.029 & - \\ 
          & \texttt{BrainFM}  &  &  \textbf{0.027} & \textbf{0.027} & \textbf{0.025} & -   \\ 
           \cmidrule(lr){2-7} 
         &  \texttt{SCRATCH} & {\multirow{3}{*}{CT}} & 0.078 & 0.055 & - & 41.42 \\
         &  \texttt{Brain-ID} &  & 0.064 & 0.047 & - & 37.90 \\ 
          & \texttt{BrainFM}  &  &  \textbf{0.053} & \textbf{0.042} & - & \textbf{34.83}   \\ 
          
\bottomrule  \\ [-3.6ex]  
    \end{tabular} 
}
\end{table}

%% file: sec/exp/fig/tab_other_tasks.tex
\begin{table}[t]
    \caption{Comparisons of \texttt{BrainFM} with \texttt{SCRATCH} and \texttt{Brain-ID}, on atlas registration (Reg), anatomy segmentation (Seg), distance prediction (Dist), and bias field estimation (BF). Corresponding train/test setups are listed in Tab.~\ref{tab: data}. Each entry reports the metric values of: (1) \texttt{L1} ($\downarrow$) for Reg; (2) \texttt{Dice} ($\uparrow$) for Seg; (3) \texttt{L1} ($\downarrow$) for Dist; and (4) \texttt{norm-L2} ($\downarrow$) for BF.} 
    \label{tab: other_tasks}
\resizebox{0.7\linewidth}{!}{
\centering 
    \begin{tabular}{ccccccc}
       \toprule \\[-3ex] 
      \multicolumn{1}{c}{\multirow{2}{*}{\bf Setup}} &  \multicolumn{1}{c}{\multirow{2}{*}{\bf Model}} & \multicolumn{1}{c}{\multirow{2}{*}{\bf \backslashbox{Input}{Output}}} & \multicolumn{1}{c}{\multirow{2}{*}{\bf Reg}} & \multicolumn{1}{c}{\multirow{2}{*}{\bf Seg}} & \multicolumn{1}{c}{\multirow{2}{*}{\bf Dist}} & \multicolumn{1}{c}{\multirow{2}{*}{\bf BF}} \\ [-0.2ex] 
           & &  \\  
     \midrule\\[-2.8ex]
       
        {\multirow{12}{*}{\texttt{\textit{I}}}} 
        & \texttt{SCRATCH} & {\multirow{3}{*}{T1w MR}} & 0.028 & 0.790 & 0.310 & 0.285 \\ 
        & \texttt{Brain-ID} &  & 0.023 & 0.822 & 0.280 & 0.270 \\ 
        & \texttt{BrainFM}  &  & \textbf{0.020} & \textbf{0.846} & \textbf{0.262} & \textbf{0.268}  \\ 
           \cmidrule(lr){2-7} 
        & \texttt{SCRATCH} & {\multirow{3}{*}{T2w MR}} & 0.025 & 0.795 & 0.320 & 0.292 \\ 
        & \texttt{Brain-ID} &  & 0.019 & 0.822 & 0.284 & 0.281 \\ 
        & \texttt{BrainFM}  &  & \textbf{0.018} & \textbf{0.832} & \textbf{0.270} & \textbf{0.279}  \\ 
           \cmidrule(lr){2-7} 
        & \texttt{SCRATCH} & {\multirow{3}{*}{FLAIR MR}} & 0.027 & 0.780 & 0.330 & 0.290 \\ 
        & \texttt{Brain-ID} &  & 0.022 & 0.810 & 0.295 & \textbf{0.275} \\ 
        & \texttt{BrainFM}  &  & \textbf{0.020} & \textbf{0.834} & \textbf{0.278} & 0.276 \\ 
           \cmidrule(lr){2-7} 
        & \texttt{SCRATCH} & {\multirow{3}{*}{CT}} & 0.034 & 0.750 & 0.340 & 0.299 \\ 
        & \texttt{Brain-ID} &  & 0.028 & 0.780 & 0.310 & 0.288 \\ 
        & \texttt{BrainFM}  &  & \textbf{0.026} & \textbf{0.802} & \textbf{0.294} & \textbf{0.286} \\ 
         \hline\\[-2.3ex]

        {\multirow{12}{*}{\texttt{\textit{II}}}} 
        & \texttt{SCRATCH} & {\multirow{3}{*}{T1w MR}} & 0.027 & 0.792 & 0.308 & 0.283 \\ 
        & \texttt{Brain-ID} &  & 0.021 & 0.828 & 0.276 & 0.268 \\ 
        & \texttt{BrainFM}  &  & \textbf{0.019} & \textbf{0.850} & \textbf{0.260} & \textbf{0.267} \\ 
           \cmidrule(lr){2-7} 
        & \texttt{SCRATCH} & {\multirow{3}{*}{T2w MR}} & 0.026 & 0.785 & 0.325 & 0.293 \\ 
        & \texttt{Brain-ID} &  & 0.020 & 0.818 & 0.288 & 0.280 \\ 
        & \texttt{BrainFM}  &  & \textbf{0.018} & \textbf{0.836} & \textbf{0.272} & \textbf{0.279} \\ 
           \cmidrule(lr){2-7} 
        & \texttt{SCRATCH} & {\multirow{3}{*}{FLAIR MR}} & 0.029 & 0.772 & 0.335 & 0.291 \\ 
        & \texttt{Brain-ID} &  & 0.024 & 0.805 & 0.300 & 0.278 \\ 
        & \texttt{BrainFM}  &  & \textbf{0.022} & \textbf{0.829} & \textbf{0.284} & \textbf{0.276} \\ 
           \cmidrule(lr){2-7} 
        & \texttt{SCRATCH} & {\multirow{3}{*}{CT}} & 0.036 & 0.740 & 0.345 & 0.301 \\ 
        & \texttt{Brain-ID} &  & 0.030 & 0.775 & 0.315 & \textbf{0.288} \\ 
        & \texttt{BrainFM}  &  & \textbf{0.027} & \textbf{0.796} & \textbf{0.298} & 0.290 \\ 
         \hline\\[-2.3ex]

        {\multirow{12}{*}{\texttt{\textit{III}}}} 
        & \texttt{SCRATCH} & {\multirow{3}{*}{T1w MR}} & 0.028 & 0.794 & 0.309 & 0.284 \\ 
        & \texttt{Brain-ID} &  & 0.022 & 0.826 & 0.278 & 0.269 \\ 
        & \texttt{BrainFM}  &  & \textbf{0.020} & \textbf{0.847} & \textbf{0.262} & \textbf{0.268} \\ 
           \cmidrule(lr){2-7} 
        & \texttt{SCRATCH} & {\multirow{3}{*}{T2w MR}} & 0.026 & 0.788 & 0.322 & 0.294 \\ 
        & \texttt{Brain-ID} &  & 0.019 & 0.820 & 0.285 & 0.282 \\ 
        & \texttt{BrainFM}  &  & \textbf{0.018} & \textbf{0.838} & \textbf{0.270} & 0.283 \\ 
           \cmidrule(lr){2-7} 
        & \texttt{SCRATCH} & {\multirow{3}{*}{FLAIR MR}} & 0.030 & 0.775 & 0.338 & 0.293 \\ 
        & \texttt{Brain-ID} &  & 0.023 & 0.808 & 0.296 & 0.279 \\ 
        & \texttt{BrainFM}  &  & \textbf{0.021} & \textbf{0.831} & \textbf{0.280} & \textbf{0.278} \\ 
           \cmidrule(lr){2-7} 
        & \texttt{SCRATCH} & {\multirow{3}{*}{CT}} & 0.037 & 0.743 & 0.348 & 0.302 \\ 
        & \texttt{Brain-ID} &  & 0.029 & 0.778 & 0.312 & \textbf{0.289} \\ 
        & \texttt{BrainFM}  &  & \textbf{0.027} & \textbf{0.800} & \textbf{0.296} & 0.290 \\ 
          
\bottomrule  \\ [-3.6ex]  
    \end{tabular} 
}
\end{table}

%% file: sec/exp/additional.tex

\subsection{Additional Experiments and Further Discussions}
\label{exp: ablation}

\input{sec/exp/fig/tab_ablat}



\subsubsection{Data Generation Design}~As shown in \cref{tab: ablat}, training with all mildly-corrupted samples results in reduced feature robustness and further harms the downstream performance, yet using all extremely corrupted samples leads to unstable training and model collapse. We train \texttt{BrainFM} on samples of gradually increased corruptions (\cref{fw: train} (left)).  

\subsubsection{Batch Size}~\texttt{BrainFM} computes training loss of all intra-subject samples in a mini-batch at once (\cref{sec: sample}). We observe that larger batches improve feature robustness~(\cref{tab: ablat}), yet do not further enhance downstream task performances.

\subsubsection{Limitations}~\texttt{BrainFM} is designed to capture the unique anatomy of subjects, it works exceptionally well on fundamental brain imaging tasks such as reconstruction, segmentation, and super-resolution, with a simple one-layer adaption. However, as \texttt{BrainFM}'s data generation is solely based on anatomies, we observe that it is not adept at handling images with extensive pathology regions. Our future work will focus on contrast-agnostic, pathology-encoded representations.

%% file: sec/exp/fig/tab_ablat.tex
\begin{table}[t]
    \caption{Comparison between \texttt{BrainFM} and its variants.} 
    \label{tab: ablat}
\resizebox{1\linewidth}{!}{
\centering 
    \begin{tabular}{lccccc}
       \toprule \\[-3ex] 
      \multicolumn{1}{c}{\multirow{2}{*}{\thead{\normalsize\textbf{Model Setup}\\\textit{\normalsize[Comparison Target]}}}} & \multicolumn{2}{c}{\textbf{Feature Robustness (Intra)}} & \multicolumn{3}{c}{\textbf{Downstream (Reconstruction)}} \\ [-0.7ex]
        \cmidrule(lr){2-3}
        \cmidrule(lr){4-6}
           & \texttt{SSIM} ($\uparrow$) & \texttt{MS-SSIM} ($\uparrow$) & \texttt{L1} ($\downarrow$) & \texttt{PSNR} ($\uparrow$) & \texttt{SSIM} ($\uparrow$) \\ [-0.2ex]
     \midrule\\[-3ex] 
         
         \multicolumn{5}{l}{\textit{[Data generation design: corruption levels in intra-subject mini-batches]}} & \\ 
        {All Mild ($\sigma_{\text{noise}}=1,\, \cdots$)} & 0.792 & 0.813 & 0.028 & 28.30 & 0.960 \\ 
        {All Medium ($\sigma_{\text{noise}}=5,\, \cdots$)} & 0.831 & 0.899 & 0.022 & 29.67 & 0.964 \\ 
        {All Severe ($\sigma_{\text{noise}}=10,\, \cdots$)} & N/A & N/A & N/A & N/A & N/A \\ 
        {Mild to Severe (\texttt{BrainFM})} & \textbf{0.858} & \textbf{0.921} & \textbf{0.021} & \textbf{29.89} & \textbf{0.966} \\ 
         \hline\\[-2.3ex]
         
        \multicolumn{5}{l}{\textit{[Mini-batch size: number of intra-subject samples]}} & \\ 
        2 & 0.826 & 0.897 & 0.025 & 28.83 & 0.959 \\ 
        3 & 0.841 & 0.902 & 0.022 & 29.30 & 0.962 \\ 
        {4~(\texttt{BrainFM})} & 0.858 & 0.921 & \textbf{0.021} & 29.89 & \textbf{0.966} \\ 
        5 & \textbf{0.860} & \textbf{0.929} & \textbf{0.021} & \textbf{29.93} & 0.965 \\ [-0.5ex]
\bottomrule  \\ [-3.6ex]  
    \end{tabular} 
}
\end{table}

%% file: sec/con.tex
\section{Conclusion}
\label{sec: con}
We introduced \texttt{BrainFM}, a multi-task foundation model for human brain imaging, which is resilient to variations in image modalities and appearances. \texttt{BrainFM} is trained on a mix-up of synthetic and real data, and can efficiently adapt to a wide array of downstream tasks. We validated \texttt{BrainFM}'s effectiveness on eight upstream and four downstream applications, expanding across eleven public datasets with both MR and CT. \texttt{BrainFM} achieves superior performances across all tasks and modalities. We believe \texttt{BrainFM} marks the first step of modality-agnostic foundation models for neuroimaging.